\documentclass[runningheads]{llncs}

\usepackage{eccv}

\usepackage{eccvabbrv}

\usepackage{graphicx}
\usepackage{booktabs}

\usepackage[accsupp]{axessibility}  
\usepackage{algorithm}
\usepackage{algorithmic}
\usepackage{amsmath}
\usepackage{svg}
\usepackage{multirow}
\usepackage{makecell}       
\usepackage{booktabs}       
\usepackage{amsfonts} 

\usepackage[breaklinks,colorlinks,citecolor=eccvblue]{hyperref}

\begin{document}


\title{BATS: Resource-Efficient Volumetric Segmentation with Boundary-Aware Mixed-Resolution Tokens}
\titlerunning{BATS: Boundary-Aware Mixed-Resolution Tokens}

%

\author{David Hagerman, Roman Naeem, Fredrik Kahl\\
\texttt{\{david.hagerman, nroman, fredrik.kahl\}@chalmers.se} \\
}

\authorrunning{D.~Hagerman et al.}

\institute{Chalmers University of Technology, 412 96 Gothenburg, Sweden}

\maketitle

\begin{abstract}
Many high-performing volumetric segmentation models maintain dense multi-scale feature maps, leading to high activation memory and inference cost. We present BATS (Boundary-Aware Token Selection), a 3D medical image segmentation architecture that concentrates fine-resolution processing near predicted class boundaries. A dense boundary predictor identifies where additional resolution is needed, while a fine-first context cascade constructs an input-dependent mixed-resolution hierarchy. Homogeneous regions are represented coarsely, with finer tokens retained around boundaries, thin structures, and small targets. The sparse hierarchy is refined and rasterised into a dense segmentation.

BATS predicts boundary relevance independently at every resolution level, preventing an erroneous coarse-scale decision from suppressing fine-scale evidence. Parent cluster attention further injects hierarchical ancestor tokens into local attention neighbourhoods, providing cross-scale context without dense multi-scale feature maps or cross-scale neighbour search.

We evaluate BATS on five public CT and MRI datasets using the standardised nnU-Net Revisited protocol. BATS achieves the highest LiTS Dice among the compared methods and averages within 0.37 Dice points of the strongest dense baseline, MedNeXt-L, across the five datasets. Relative to MedNeXt-L, it reduces peak allocated GPU memory by more than 53\% on KiTS, LiTS, and BraTS. Inference is up to 30\% faster on KiTS and LiTS, which retain fewer tokens, but slower on the more token-dense BraTS. Mixed-resolution processing therefore provides consistent memory savings, while runtime and accuracy gains depend on dataset boundary density.
\end{abstract}

\section{Introduction}
\label{sec:intro}

Many high-performing volumetric medical image segmentation networks process dense multi-scale feature grids throughout every input crop. This holds across CNN-based models such as nnU-Net~\cite{isensee_nnu-net_2021} and MedNeXt~\cite{roy2023mednext} and transformer-based architectures such as SwinUNETR~\cite{tang_self-supervised_2022}, nnFormer~\cite{zhou_nnformer_2022}, and UNETR++~\cite{shaker2024unetr++}: although their building blocks differ, each region follows essentially the same resolution schedule. This is expensive in 3D because the number of feature locations grows cubically with spatial resolution, making activation memory and inference time primary constraints.

The required spatial detail, however, is not uniform. Boundaries, small lesions, and thin anatomical structures require fine representation, whereas homogeneous tissue and background can often be represented more coarsely without losing relevant semantic information. This motivates densely analysing each input crop for localisation while restricting expensive fine-resolution refinement to regions whose predicted structure requires it.

Adaptive token methods reduce computation by pruning, merging, halting, or routing tokens~\cite{rao2021dynamicvit,yin2022vit,bolya2022token,ziwen_autofocusformer_2023}. Most use a fixed retention schedule or token budget, adapting which tokens are processed but not the amount of computation allocated to each individual input. This distinction matters in medical imaging, where the number, size, and geometry of targets vary substantially between scans.

ARTA~\cite{hagerman2026arta} introduced boundary-guided adaptive-resolution tokenisation for 2D semantic segmentation, allowing both the number and spatial distribution of fine tokens to depend on the input. Extending this approach to 3D introduces three challenges. First, iterative coarse-to-fine boundary prediction with hard gating can irreversibly discard small structures when they are missed at a coarse level. Second, a sparse mixed-resolution hierarchy needs cross-scale interaction without creating dense multi-scale feature maps. Third, token sparsity must translate into end-to-end reductions in memory and runtime rather than merely reducing an intermediate token count.

We address these requirements with BATS, which combines a dense boundary predictor, a sparse mixed-resolution refiner, and a lightweight segmentation head. The predictor scores all split levels densely and in parallel, after which a fine-first context cascade retains selected tokens and their ancestor chains. This recall-oriented design prevents a coarse false negative from suppressing valid fine-scale evidence. Parent cluster attention then injects retained ancestors into each token's local attention neighbourhood, guaranteeing spatially aligned context from every coarser scale without reconstructing dense multi-scale refiner features. The resulting hierarchy varies in both token count and spatial distribution across inputs before being rasterised into voxel-level logits.

We evaluate BATS on ACDC, LiTS, BraTS, KiTS, and AMOS under the standardised nnU-Net Revisited protocol [12]. BATS achieves the highest LiTS Dice and performs within 0.37 Dice points of MedNeXt-L on average across the five datasets. Under matched inference settings, it uses less than half the peak allocated GPU memory of MedNeXt-L on KiTS, LiTS, and BraTS. Runtime decreases on the relatively sparse KiTS and LiTS datasets but increases on BraTS, where the retained token hierarchy is substantially denser. These results establish BATS as a memory-efficient alternative with a dataset-dependent runtime–accuracy trade-off.

Our main contributions are:
\begin{itemize}
    \item BATS, a resource-efficient volumetric segmentation architecture that combines dense boundary analysis with input-dependent sparse refinement, allowing both token count and spatial resolution to adapt to each input.

    \item Parallel boundary prediction and a fine-first context cascade that avoid hard coarse gating and the resulting irreversible fine-scale misses of iterative coarse-to-fine methods.

    \item Parent cluster attention, which supplies sparse tokens with deterministic hierarchical context without dense multi-scale refiner features or token-level nearest-neighbour search.

    \item An evaluation under the nnU-Net Revisited protocol demonstrating more than 53\% lower peak allocated GPU memory than MedNeXt-L on three datasets, with runtime gains on datasets that produce sparse token hierarchies.
\end{itemize}

\section{Related Work}
\label{sec:rel_work}

\subsection{3D medical image segmentation}

Dense CNN encoders remain strong baselines for volumetric medical segmentation. nnU-Net~\cite{isensee_nnu-net_2021} automatically configures a U-Net-style architecture from dataset properties and remains a strong reference across tasks. MedNeXt~\cite{roy2023mednext} adapts ConvNeXt to 3D using large depthwise kernels, achieving strong results with relatively few parameters. Transformer-based architectures have also been extended to volumetric data. SwinUNETR~\cite{tang_self-supervised_2022} uses shifted-window self-attention in a 3D U-Net framework and nnFormer~\cite{zhou_nnformer_2022} combines local and global self-attention.

These architectures process volumes on dense grids and apply the same resolution schedule to homogeneous tissue, object boundaries, and small target structures alike. BATS instead uses boundary predictions to construct a sparse mixed-resolution representation whose density varies across both space and samples.

\subsection{Boundary- and saliency-guided medical segmentation}

A related line of work learns where to focus within dense architectures. Attention U-Net~\cite{oktay2018attention} and Attention Gated Networks~\cite{schlemper2019attention} use learned saliency to reweight dense features and provide implicit localisation, while BATFormer~\cite{batformer} adapts local transformer windows using entropy-guided, boundary-related cues. These methods adapt feature weighting or attention support while retaining dense spatial feature maps throughout the network. BATS instead uses predicted class transitions to determine the spatial resolution of the refiner input itself: which regions are represented at fine resolution and which remain coarse.

\subsection{Adaptive token selection and sparse computation}

General vision methods reduce token computation through learned pruning, adaptive halting, or token merging, including DynamicViT~\cite{rao2021dynamicvit}, A-ViT~\cite{yin2022vit}, and ToMe~\cite{bolya2022token}. AutoFocusFormer~\cite{ziwen_autofocusformer_2023} constructs spatial token clusters across hierarchical stages but uses a predefined retention rate and a single token resolution within each stage. These methods demonstrate that dense token grids contain substantial redundancy, but their pruning schedule, retention ratio, or token budget is typically fixed at design time, so they adapt the identity of the retained tokens rather than the amount of computation required by each input.

The limitation of a fixed token budget is particularly consequential in volumetric medical segmentation. A scan containing one small lesion requires fine resolution in different locations and quantities than a scan containing multiple lesions or thin vessels, and neither should share the fine-token budget of a scan dominated by homogeneous anatomy. ARTA~\cite{hagerman2026arta} addresses this in 2D semantic segmentation by making both the number and placement of fine-resolution tokens input-dependent, driven by predicted class-boundary structure. BATS applies the same principle in 3D, redesigning boundary prediction, token-hierarchy construction, and cross-scale attention for the scale and sparsity of volumetric data.

Recent medical segmentation methods have also explored sparse token processing. APFormer~\cite{lin2023lighter} applies adaptive transformer pruning to medical image segmentation. Zhou et al.~\cite{zhou2023token} formulate medical segmentation as sparse encoding followed by token completion and dense decoding, using Soft-topK token pruning to select a fixed number of tokens. HRViT~\cite{guo2025edge} performs edge-aware token halting for 3D medical segmentation, allowing edge tokens to continue to deeper layers while other tokens are halted. TokenSeg~\cite{zeng2026tokenseg} proposes boundary-aware sparse token compression for 3D medical segmentation by selecting a fixed number of salient tokens from multi-scale candidates.

These methods differ in whether token count is fixed or input-dependent, whether multiple resolutions coexist within the active set, and whether token selection is sequential or parallel. Most sparse medical segmentation methods use predefined pruning or compression budgets. AutoFocusFormer uses fixed retention rates and one active resolution per stage, whereas BATS jointly refines a variable-size mixed-resolution hierarchy. ARTA makes token count input-dependent but constructs the token hierarchy iteratively, so a coarse miss prevents finer evaluation. BATS instead scores all split levels in parallel and uses explicit ancestor links for cross-scale interaction. Because the published implementations of ARTA, DynamicViT, ToMe, and AutoFocusFormer are not directly applicable to the present 3D medical setting, we compare their token-selection principles conceptually rather than numerically.

\subsection{Local and cross-scale attention}

Efficient attention mechanisms restrict the set of tokens each query attends to. The Swin Transformer~\cite{swin} partitions the image into local windows and alternates shifted window partitions to exchange information across windows. Cluster Attention~\cite{ziwen_autofocusformer_2023} instead uses clusters to create local neighbourhoods around each token. These mechanisms reduce attention cost, but local attention alone provides limited interaction between resolutions in a sparse mixed-scale token set.

Cross-scale attention is commonly handled using dense multi-scale feature maps or spatial neighbour search. Multi-Scale Deformable Attention~\cite{zhu_deformable_2020} attends to learned offset locations across dense feature maps, while point-cloud methods often rely on $k$-nearest-neighbour search in 3D space~\cite{ziwen_autofocusformer_2023}. BATS instead exploits deterministic parent links in the token hierarchy. Ancestors are retrieved through direct hash-grid lookup and injected into each token's attention neighbourhood, providing fine-resolution tokens with guaranteed access to coarser context without dense feature maps or token-level nearest-neighbour search.

\section{Method}
\label{sec:method}

\subsection{Architecture overview}
\label{sec:overview}

\begin{figure*}[!ht]
  \begin{center}
  \includegraphics[width=0.95\textwidth]{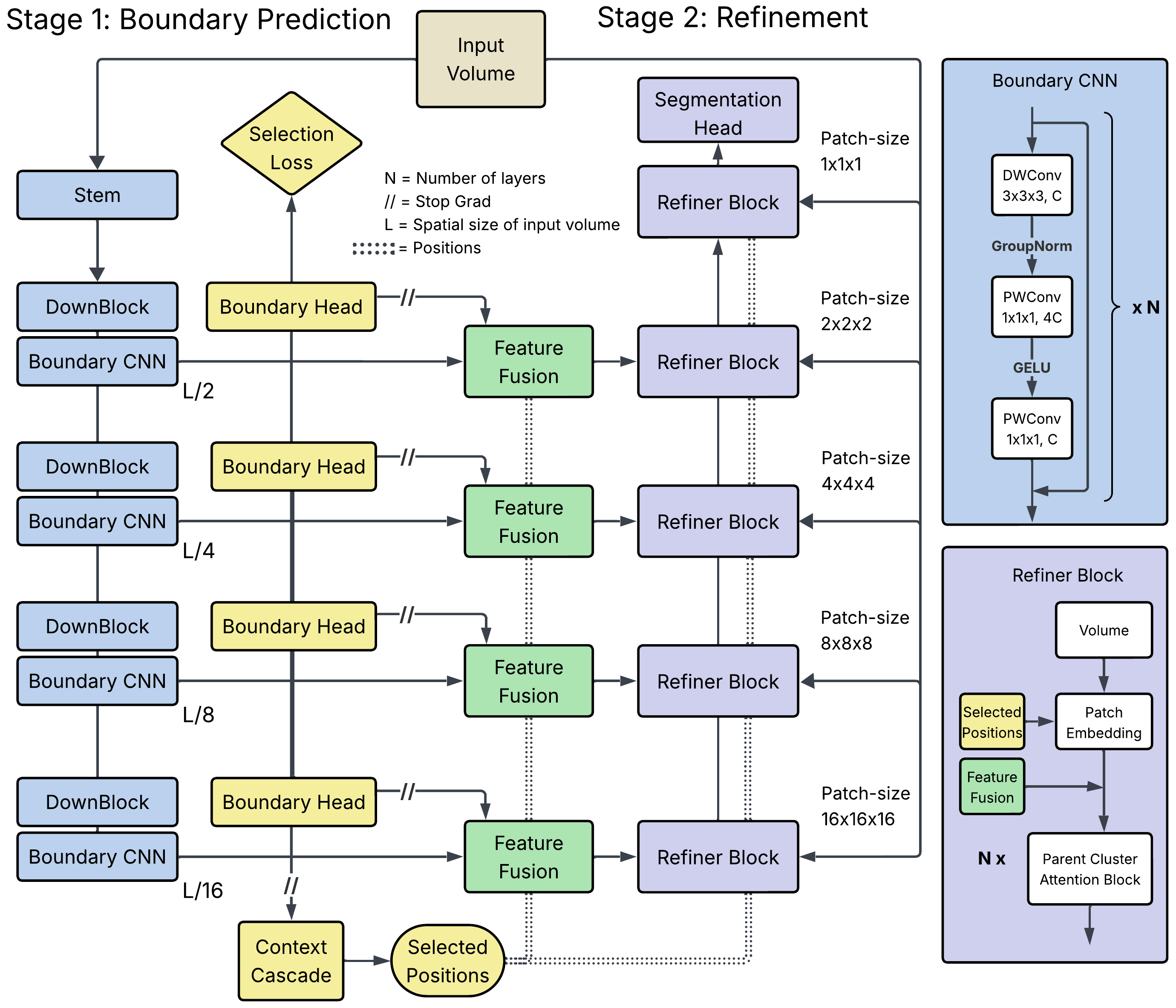}
  \caption{BATS architecture. A dense CNN predicts split scores at the four patch scales ($16^3$, $8^3$, $4^3$, $2^3$). The fine-first context cascade converts these scores into a mixed-resolution token hierarchy and retains each selected token's ancestors. The refiner combines FPN and raw-patch features and processes the hierarchy using parent cluster attention, which augments each token's local neighbourhood with injected ancestor tokens. A dense head rasterises the refined tokens into voxel-level logits.}
  \label{fig:bats_architecture}
  \end{center}
\end{figure*}

BATS represents an input crop using a hierarchy of $N_s=5$ token scales with cubic patch sizes $p_s^3$, where $p_s \in \{16,8,4,2,1\}$ from coarsest to finest. All tokens at the coarsest patch scale, $16^3$, are retained, while finer tokens are introduced according to predicted split decisions. A dense boundary predictor (Section~\ref{sec:boundary_predictor}) scores the four split levels $\{16^3,8^3,4^3,2^3\}$ in parallel, and a fine-first context cascade converts these scores into a sparse hierarchy containing each selected token and its ancestors. A refiner (Section~\ref{sec:refiner}) processes the hierarchy using parent cluster attention, after which a lightweight dense head (Section~\ref{sec:seg_head}) rasterises the token features into voxel-level logits. Figure~\ref{fig:bats_architecture} shows the architecture; Table~\ref{tab:architecture} summarises the main configuration and the full stage settings are given in Appendix~\ref{app:impl}.

\begin{table}[!ht]
\centering
\caption{Main BATS configuration: token width and block count per scale for the boundary predictor ($19$M) and refiner ($34$M); the dense segmentation head adds $0.4$M, for $53$M in total. Full stage settings (heads, cluster and neighbourhood sizes, expansion ratios, regularisation) are listed in Appendix~\ref{app:impl}.}
\label{tab:architecture}
\small
\setlength{\tabcolsep}{6pt}
\begin{tabular}{ll rrrrr}
\toprule
Component & Scale & $16^3$ & $8^3$ & $4^3$ & $2^3$ & $1^3$ \\
\midrule
\multirow{2}{*}{Boundary predictor} & Width  & 512 & 256 & 128 & 64 & -- \\
                                    & Blocks & 6 & 8 & 4 & 3 & -- \\
\midrule
\multirow{2}{*}{Refiner}            & Width  & 512 & 256 & 128 & 64 & 32 \\
                                    & Blocks & 8 & 8 & 8 & 4 & 2 \\
\bottomrule
\end{tabular}
\end{table}

\subsection{Boundary prediction and token selection}
\label{sec:boundary_predictor}

Because the locations of small lesions and thin structures cannot be assumed a priori, the boundary predictor must evaluate potential boundary regions densely. It must also produce predictions at all split levels without gating by coarser decisions, so that fine-scale evidence can recover regions that appear uniform at coarse resolution. We therefore use a 3D ConvNeXt pyramid with an FPN~\cite{roy2023mednext} that predicts split logits at patch sizes $16^3$, $8^3$, $4^3$, and $2^3$, with one head per scale evaluated densely in one forward pass. Adjacent split maps are connected by a learned gated residual, allowing coarse evidence to reinforce finer estimates without conditioning any fine-level output on a discrete coarser decision. Block, FPN, and gate details are given in Appendix~\ref{app:impl}.

\paragraph{Fine-first context cascade.}
A patch is split when its boundary probability exceeds $\tau{=}0.5$, and each positive decision introduces its children at the next finer token scale. We apply split decisions from fine to coarse and add the complete ancestor chain of every selected token. Consequently, fine-scale evidence is retained even when the corresponding coarse prediction is negative, while every selected token remains connected to the full hierarchy used by the refiner. After training, the split heads typically produce strongly bimodal scores, making the resulting token hierarchies relatively stable for thresholds near $\tau=0.5$.

\paragraph{Training objective and oracle warmup.}
Each split head is supervised with a Dice-BCE loss against a binary target that is $1$ where a patch contains more than one semantic class, i.e.\ $\max_{\text{patch}}(\text{labels}) \neq \min_{\text{patch}}(\text{labels})$, and $0$ otherwise. Early in training, rapidly changing predicted hierarchies destabilise the refiner input. At each optimisation step, we therefore use the ground-truth hierarchy for the entire batch with probability $q(t)$. $q(t)$ decays linearly from $0.8$ to $0$ over the first $25\%$ of training and is $0$ thereafter.

\subsection{Sparse refinement and parent cluster attention}
\label{sec:refiner}

Each retained token combines two signals: an embedding of its raw image patch and the boundary-predictor FPN feature at the corresponding position. A per-scale patch encoder reduces the raw $p_s^3$ patch to one vector, while the gathered FPN feature is projected to the model width. We fuse them as
\begin{equation}
    \mathbf{f}_s = g_s\, \mathbf{f}^{\text{img}}_s + w_s\, \mathbf{f}^{\text{FPN}}_s,
\end{equation}
where $g_s=\sigma(\gamma_s)$ is a learned scale-specific fusion gate and $w_s=\sigma(\ell_{s-1})$ is the split probability of the token's parent patch; we set $w_s=1$ at the coarsest scale, which has no parent. Voxel tokens have no FPN level and use only the image embedding. The gate is initialised to favour the FPN path, and gradients flow through both branches into the boundary predictor; further fusion details are given in Appendix~\ref{app:impl}.

Refinement proceeds from coarse to fine. The $16^3$ tokens are processed first; at each subsequent stage, the next finer tokens are introduced and the full active hierarchy is processed jointly. Coarse tokens therefore already contain contextual information when supplied as ancestors to finer tokens.

\begin{figure*}[!ht]
  \begin{center}
  \includegraphics[width=0.95\textwidth]{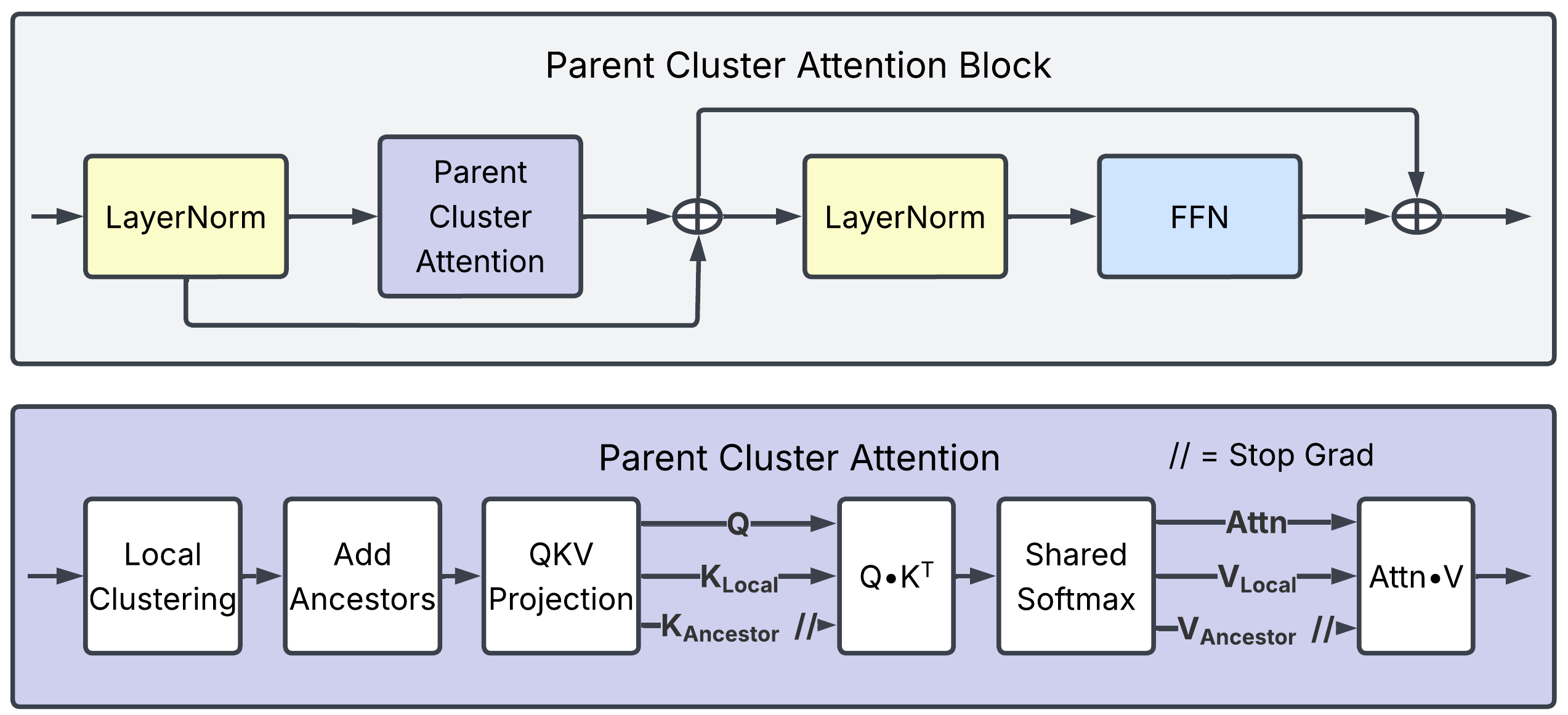}
  \caption{Parent cluster attention block and mechanism. Retained ancestors augment the local neighbourhood, with detached ancestor keys contributing to the shared attention softmax and the corresponding values to the weighted aggregation.}
  \label{fig:parent_cluster_attention}
  \end{center}
\end{figure*}

\paragraph{Parent cluster attention.}
\label{sec:parent_attn}
In a sparse mixed-resolution 3D hierarchy, the number of fine tokens grows cubically with resolution, causing them to dominate local spatial neighbourhoods and making cross-scale interactions rare under ordinary local attention. Yet a fine token's ancestors represent the same location at coarser scales and carry spatially aligned context that helps disambiguate its class. Parent cluster attention makes this dependency explicit by injecting ancestor tokens directly into each token's attention set. See Figure~\ref{fig:parent_cluster_attention} for an overview.

We adapt the balanced clustering of AutoFocusFormer~\cite{ziwen_autofocusformer_2023} to the mixed-resolution setting. Every token is represented by its patch centre in units of the finest patch size, so tokens of different scales covering the same region occupy nearby coordinates. Tokens are ordered spatially and grouped into clusters of $m = 8$ (derivation of the ordering follows~\cite{ziwen_autofocusformer_2023}). For each query token $i$, the tokens belonging to the $R$ clusters with centroids nearest to $\mathbf{p}_i$ form the local neighbourhood $\mathcal{N}_{\mathrm{local}}(i)$. Nearest-neighbour search is therefore performed only over cluster centroids rather than over all tokens, while overlapping neighbourhoods permit information exchange between clusters. The resulting neighbourhood contains between 48 tokens at the coarsest stage and 16 tokens at the finest stage, as detailed in Appendix~\ref{app:impl}.

Local clustering alone does not guarantee interaction between resolution levels. For a token $i$ with origin $(x,y,z)$ and patch size $p_s$, the ancestor at a coarser patch size $p_c>p_s$ has origin $\left(p_c\left\lfloor x/p_c\right\rfloor,p_c\left\lfloor y/p_c\right\rfloor,p_c\left\lfloor z/p_c\right\rfloor\right)$. Let $\mathcal{A}(i)$ be the set of retained ancestors of $i$, one per coarser scale. Parent cluster attention uses
\begin{equation}
    \mathcal{N}_{\text{parent}}(i) = \mathcal{N}_{\text{local}}(i) \cup \mathcal{A}(i)
\end{equation}
as the key/value set for query $i$. Ancestor indices follow directly from the hierarchical patch coordinates and are retrieved from a hash grid indexed by $(\mathrm{batch},\mathrm{scale},x,y,z)$, eliminating token-level search. The fine-first context cascade guarantees that all required ancestors exist for every non-coarsest token. Queries, local neighbours, and ancestors share a single softmax, with RoPE3D encoding their spatial positions and a learned scale embedding distinguishing resolution levels.

Because many fine tokens share the same ancestors, accumulating gradients through the injected ancestor keys and values causes substantial write contention. We therefore stop gradients through these injected copies. The ancestor tokens remain trainable through their roles as queries and local neighbours, and the forward attention computation is unchanged. Further details on batching and masking are given in Appendix~\ref{app:impl}.

\subsection{Segmentation head and training objectives}
\label{sec:seg_head}
\label{sec:deep_supervision}

Refined tokens at each scale are projected to a shared output width by a two-layer MLP. Each voxel is assigned the feature of the finest retained token covering its location, thereby rasterising the sparse hierarchy into a dense feature volume. A shallow residual CNN combines this volume with a convolutional image skip, computed from the raw input using a stride equal to the finest patch size, and predicts the final segmentation logits. The boundary predictor and fine-first context cascade therefore determine where fine-resolution representation is required, while the refiner processes the resulting sparse hierarchy without reconstructing dense multi-scale feature maps. 

The training objective combines the final Dice-CE segmentation loss with the Dice-BCE boundary-prediction loss and two auxiliary objectives. Both auxiliary heads predict the class-fraction distribution of the corresponding patch using soft cross-entropy. The first operates on boundary-predictor features at each split level, while the second operates on intermediate refiner tokens at every scale. Each auxiliary loss receives weight $0.15$, and both heads are discarded at inference. Their architectures are described in Appendix~\ref{app:impl}, and their contributions are evaluated in the component ablation in Table~6.

\section{Results}

\subsection{Experimental setup}
We evaluate BATS on five datasets from the nnU-Net Revisited benchmark~\cite{isensee2024nnu}: ACDC~\cite{bernard2018deep} (cardiac cine-MRI; $n=200$), LiTS~\cite{bilic2023liver} (liver and liver-tumour CT; $n=131$), BraTS~\cite{baid2021rsna, menze2014multimodal} (multi-modal brain-tumour MRI; $n=1251$), KiTS~\cite{heller2023kits21} (kidney, tumour, and cyst CT; $n=489$), and AMOS~\cite{ji2022amos} (multi-organ abdominal CT; $n=360$). Because differences in preprocessing, data splits, and training budgets often confound comparisons in 3D medical segmentation, we adopt the standardised nnU-Net Revisited protocol and take all baseline results directly from that benchmark. Table~\ref{tab:nnunet_revisited_results} reports five-fold cross-validation using exactly the same split folds as nnU-Net and MedNeXt. BTCV~\cite{landman2015btcv} is omitted because its 30 cases yield only six validation cases per fold, making the fold-level results highly sensitive to individual cases.

We use the benchmark's preprocessing, augmentation, training duration, and sliding-window protocol, taken unchanged from MedNeXt~\cite{roy2023mednext,isensee2024nnu}. BATS is trained on $128^3$ crops with batch size $2$ using AdamW, Dice-CE segmentation loss, and Dice-BCE boundary loss, with a general learning rate of $10^{-3}$ and weight decay $10^{-2}$ for the refiner and $10^{-4}$ for the boundary predictor; the two auxiliary losses receive weight $0.15$ each. Inference uses sliding-window evaluation with $0.5$ overlap, Gaussian aggregation, and FP16. Dice is computed per class over the validation cases; for KiTS and BraTS, whose targets form nested regions, we report Dice over the hierarchical evaluation classes (HEC) specified by the respective challenges. Remaining optimisation and preprocessing settings are given in Appendix~\ref{app:impl}.

\subsection{Accuracy and efficiency}

\subsubsection{Benchmark comparison.}

\begin{table*}[!ht]
\centering
\caption{Benchmark results for methods evaluated under the nnU-Net Revisited protocol~\cite{isensee2024nnu}.
DSC scores are reported in percent; the best value per column is in bold. Parameter counts and baseline scores are taken from nnU-Net Revisited.}
\label{tab:nnunet_revisited_results}
\small
\setlength{\tabcolsep}{4pt}
\renewcommand{\arraystretch}{1.05}

\centering
\begin{tabular}{l*{6}{r}}
\toprule
Method
& \makecell{Params\\(M)}
& ACDC
& LiTS
& BraTS
& KiTS
& AMOS \\
\midrule

nnU-Net \cite{isensee_nnu-net_2021}
& 31 & 91.54 & 80.09 & 91.24 & 86.04 & 88.64 \\

nnU-Net ResEnc M \cite{isensee_nnu-net_2021}
& 31 & 91.99 & 80.75 & 91.26 & 86.79 & 88.77 \\

nnU-Net ResEnc L \cite{isensee_nnu-net_2021}
& 102 & 91.69 & 81.60 & 91.13 & 88.17 & 89.41 \\

nnU-Net ResEnc XL \cite{isensee_nnu-net_2021}
& 140 & 91.48 & 81.19 & 91.18 & \textbf{88.67} & 89.68 \\

\midrule

MedNeXt-L k3 \cite{roy2023mednext}
& 62 & \textbf{92.65} & 82.14 & 91.35 & 88.25 & 89.62 \\

MedNeXt-L k5 \cite{roy2023mednext}
& 63 & 92.62 & 82.34 & \textbf{91.50} & 87.74 & \textbf{89.73} \\

\midrule

STU-Net S \cite{huang2023stu}
& 15 & 91.04 & 78.50 & 90.55 & 84.93 & 88.08 \\

STU-Net B \cite{huang2023stu}
& 58 & 91.30 & 79.19 & 90.85 & 86.32 & 88.46 \\

STU-Net L \cite{huang2023stu}
& 440 & 91.31 & 80.31 & 91.26 & 85.84 & 89.34 \\

\midrule

SwinUNETR \cite{tang_self-supervised_2022}
& 63 & 91.29 & 76.50 & 90.68 & 81.27 & 83.81 \\

SwinUNETRV2 \cite{he2023swinunetr}
& 73 & 92.01 & 77.85 & 90.74 & 84.14 & 86.24 \\

nnFormer \cite{zhou_nnformer_2022}
& 150.5 & 92.40 & 77.40 & 90.22 & 75.85 & 81.55 \\

CoTr \cite{xie2021cotr}
& 50 & 90.56 & 79.10 & 90.73 & 84.59 & 88.02 \\

\midrule

U-Mamba Bot \cite{ma2024u}
& 42 & 91.79 & 80.40 & 91.26 & 86.22 & 89.13 \\

U-Mamba Enc \cite{ma2024u}
& 43 & 91.22 & 80.27 & 90.91 & 86.34 & 88.38 \\

\midrule

A3DS SegResNet \cite{myronenko20183d, myronenko2024auto3dseg}
& 87 & 90.69 & 79.28 & 90.79 & 81.11 & 87.27 \\

A3DS DiNTS \cite{He_2021_CVPR, myronenko2024auto3dseg}
& 152 & 82.97 & 69.05 & 87.75 & 65.28 & 82.35 \\

A3DS SwinUNETR \cite{tang_self-supervised_2022, myronenko2024auto3dseg}
& 63 & 82.68 & 68.59 & 89.90 & 52.82 & 85.05 \\

\midrule

BATS (ours)
& 53 & 92.16 & \textbf{83.39} & 91.11 & 87.02 & 88.50 \\

\bottomrule
\end{tabular}
\end{table*}

BATS achieves the highest LiTS score in Table~2 at 83.39 Dice, outperforming MedNeXt-L k3 by 1.25 points. Relative to the same baseline, BATS is lower by 0.49, 0.24, 1.23, and 1.12 points on ACDC, BraTS, KiTS, and AMOS, respectively, resulting in an average difference of $-0.37$ points across the five datasets. BATS uses 53M parameters, compared with 62M for MedNeXt-L k3.

\subsubsection{Retained token hierarchy.}
Table~3 reports per-case token counts, obtained by first averaging over each case's sliding-window crops and then computing the mean and standard deviation across cases. Across all five datasets, the retained hierarchy contains only a small fraction of the $128^3=2{,}097{,}152$ tokens required by a fully resolved voxel-level representation. The coarsest $16^3$ patch scale contributes a fixed 512 tokens per crop, while most retained tokens belong to the $2^3$ and $1^3$ scales. KiTS produces the sparsest hierarchy, with 9.3k tokens per crop on average, whereas BraTS produces the densest at 52.3k, approximately five times as many. The standard deviations are large relative to the means, demonstrating substantial case-to-case variation in the token count produced by boundary prediction.

The retained hierarchy therefore reflects the amount and distribution of predicted boundary structure rather than volume size alone. Foreground-biased KiTS and LiTS crops typically contain a single organ surrounded by large homogeneous or background regions, resulting in relatively few fine-resolution tokens. BraTS crops contain extensive brain boundaries and tumour-subregion interfaces, producing many class transitions and approximately five times as many retained tokens as KiTS. The relationship between token density and runtime is examined next.

\begin{table}[!ht]
\centering
\caption{Retained tokens per $128^3$ crop, reported as mean $\pm$ standard deviation across cases. Counts are in thousands of tokens; the fixed count of 0.512 at the $16^3$ scale corresponds to 512 tokens.}
\label{tab:token_stats}
\footnotesize
\setlength{\tabcolsep}{3pt}
\begin{tabular}{lcccccc}
\toprule
& \multicolumn{6}{c}{Tokens ($\times 10^3$)} \\
\cmidrule(lr){2-7}
Dataset & Total & $16^3$ & $8^3$ & $4^3$ & $2^3$ & $1^3$ \\
\midrule
ACDC
& $21.3 \pm 8.8$
& $0.512$
& $0.52 \pm 0.06$
& $1.4 \pm 0.4$
& $5.1 \pm 2.0$
& $13.7 \pm 6.3$ \\

LiTS
& $19.5 \pm 7.6$
& $0.512$
& $0.54 \pm 0.07$
& $1.2 \pm 0.4$
& $4.0 \pm 1.7$
& $13.2 \pm 5.4$ \\

BraTS
& $52.3 \pm 26.8$
& $0.512$
& $0.57 \pm 0.18$
& $2.4 \pm 1.2$
& $11.1 \pm 5.7$
& $37.7 \pm 19.8$ \\

KiTS
& $9.3 \pm 4.4$
& $0.512$
& $0.44 \pm 0.02$
& $0.7 \pm 0.3$
& $2.1 \pm 1.0$
& $5.6 \pm 3.1$ \\

AMOS
& $33.9 \pm 21.9$
& $0.512$
& $0.66 \pm 0.18$
& $2.1 \pm 1.1$
& $7.7 \pm 4.7$
& $23.0 \pm 15.9$ \\
\bottomrule
\end{tabular}
\end{table}

\subsubsection{Efficiency.}

We focus on inference efficiency because full-volume sliding-window evaluation produces a different token distribution from the foreground-biased crops sampled during training. Training runtime is included as a secondary comparison. We compare against MedNeXt-L k3 and k5, the strongest dense baselines of comparable parameter count ($62$--$63$M versus 53M). Peak allocated GPU memory and per-volume inference time are measured on KiTS, LiTS, and BraTS using identical settings for all methods: batch size 1, 0.5 window overlap, Gaussian aggregation, and FP16 on the same NVIDIA A100 40GB GPU. The first case is excluded because MedNeXt performs a cuDNN benchmarking pass that transiently increases memory use and runtime; it therefore serves as a warm-up for all methods. Inference time is reported as mean $\pm$ standard deviation across cases, with the large variation reflecting differences in volume size. MedNeXt training runtimes are taken from nnU-Net Revisited rather than remeasured in our environment and should therefore be treated as an approximate secondary comparison. Further measurement details are given in Appendix~\ref{app:efficiency}.

\begin{table*}[!ht]
\centering
\caption{Efficiency comparison with MedNeXt-L on KiTS, LiTS, and BraTS.
Peak memory and inference time are measured under identical settings.
MedNeXt training times are taken from nnU-Net Revisited.
Accuracy results are reported in Table~\ref{tab:nnunet_revisited_results}.}
\label{tab:efficiency}
\footnotesize
\setlength{\tabcolsep}{6pt}
\renewcommand{\arraystretch}{1.05}

\begin{tabular}{llccc}
\toprule
Dataset & Method
& \makecell{Peak mem.\\(GB)}
& \makecell{Inference\\(s/vol.)}
& \makecell{Training\\(GPU-h)} \\
\midrule

\multirow{3}{*}{KiTS}
& MedNeXt-L k3 & $2.45$ & $34.3 \pm 20.9$ & $68$ \\
& MedNeXt-L k5 & $2.46$ & $57.6 \pm 36.3$ & $233$ \\
& BATS (ours)  & $1.12$ & $23.9 \pm 11.2$ & $50$ \\
\addlinespace

\multirow{3}{*}{LiTS}
& MedNeXt-L k3 & $2.45$ & $32.3 \pm 16.6$ & $68$ \\
& MedNeXt-L k5 & $2.46$ & $54.6 \pm 28.6$ & $233$ \\
& BATS (ours)  & $1.14$ & $23.5 \pm 10.3$ & $44$ \\
\addlinespace

\multirow{3}{*}{BraTS}
& MedNeXt-L k3 & $2.48$ & $1.36 \pm 0.20$ & $68$ \\
& MedNeXt-L k5 & $2.49$ & $2.31 \pm 0.36$ & $233$ \\
& BATS (ours)  & $1.15$ & $1.75 \pm 0.36$ & $60$ \\

\bottomrule
\end{tabular}
\end{table*}

BATS uses 1.12--1.15 GB of peak allocated memory, compared with 2.45--2.49 GB for the two MedNeXt-L variants, corresponding to reductions of more than 53\% on all three datasets. Peak memory is determined by the crop producing the largest retained hierarchy rather than by the average token count. Even this worst-case crop remains substantially below the dense baselines because a considerable fraction of its spatial support is represented at coarse or intermediate resolution. The memory reduction therefore persists across datasets and baseline kernel sizes, including on BraTS despite its substantially higher average token count.

Runtime is more dataset-dependent and broadly follows the token statistics in Table~3. On KiTS and LiTS, BATS is 30.3\% and 27.2\% faster than MedNeXt-L k3, requiring 23.9 versus 34.3 seconds and 23.5 versus 32.3 seconds per volume, respectively. Relative to MedNeXt-L k5, the corresponding reductions are approximately 58\% and 57\%.

On BraTS, whose volumes are smaller but whose crops produce the densest retained hierarchies, BATS requires 1.75 seconds per volume and is approximately 29\% slower than k3 at 1.36 seconds, although it remains 24\% faster than k5 at 2.31 seconds. In this setting, the larger retained hierarchy and the fixed overhead of gathering and scattering variable-length token sets outweigh the computation saved in coarsely represented regions.

The accuracy--efficiency trade-off also varies by dataset (Table~2). On LiTS, BATS improves Dice by 1.25 points over k3 while reducing both memory and runtime, constituting a Pareto improvement. On KiTS, it trades 1.23 HEC Dice points for lower memory use and faster inference. Training runtime follows the same general ordering, with BATS requiring 50, 44, and 60 GPU-hours on KiTS, LiTS, and BraTS, respectively, compared with the externally reported 68 GPU-hours for k3 and 233 GPU-hours for k5. These results indicate that the sparsity of the retained boundary-aware hierarchy, rather than volume size alone, is a key determinant of whether BATS provides a favourable runtime trade-off.


\newcommand{\qimgw}{0.168\textwidth}
\newcommand{\qimgh}{0.168\textwidth}

\newcommand{\qboxw}{0.17\textwidth}
\newcommand{\qboxh}{0.17\textwidth}

\newcommand{\qfull}[1]{%
  \parbox[c][\qboxh][c]{\qboxw}{%
    \centering
    \includegraphics[width=\qimgw,height=\qimgh]{#1}%
  }%
}

\newcommand{\qzoom}[1]{%
  \parbox[c][\qboxh][c]{\qboxw}{%
    \centering
    \includegraphics[width=\qimgw,height=\qimgh]{#1}%
  }%
}

\subsection{Ablation studies}

\subsubsection{Token-hierarchy construction: parallel, iterative, and random.}

We compare BATS with an iterative ARTA-style boundary-prediction baseline of matched capacity and a random-placement baseline on a single KiTS fold, while holding the refiner, segmentation head, losses, and hyperparameters fixed. The iterative baseline evaluates scale $s$ only at positions retained at scale $s-1$~\cite{hagerman2026arta}. Because it operates on a progressively sparse hierarchy, it uses attention-based boundary-prediction blocks rather than BATS's dense CNN, so the experiment compares complete parallel and iterative designs at matched capacity rather than isolating an individual component. The random baseline receives BATS's per-scale token count and places the tokens uniformly at random, separating the effect of spatial placement from that of the token budget. To distinguish boundary-prediction errors from refinement errors, we additionally report precision and recall at several split-depth thresholds. For each voxel, the ground-truth and predicted split depths count the number of positive decisions along its hierarchy; threshold $k$ requires at least $k$ splits, with $k=1$ and $k=4$ corresponding to the coarsest and finest split decisions, respectively. Appendix~\ref{app:selection_metrics} provides the formal definition.

\begin{table*}[!ht]
\centering
\caption{Token-hierarchy construction strategies on KiTS at matched boundary-prediction parameter count. HEC Dice is averaged over the hierarchical evaluation classes. Split-depth recall and precision are reported at thresholds $k=1,\ldots,4$. Memory and inference time are measured as in Table~\ref{tab:efficiency}.}
\label{tab:ablation_selection}
\small
\setlength{\tabcolsep}{3pt}
\renewcommand{\arraystretch}{1.05}

\begin{tabular}{lccc c cccc}
\toprule
Model
& \makecell{HEC\\Dice}
& \makecell{Mem.\\(GB)}
& \makecell{Inference\\(s/vol.)}
& Metric
& $k{=}1$
& $k{=}2$
& $k{=}3$
& $k{=}4$ \\
\midrule

\multirow{2}{*}{Iterative}
& \multirow{2}{*}{$84.55$}
& \multirow{2}{*}{$1.21$}
& \multirow{2}{*}{$30.2 \pm 14.3$}
& Recall (\%)    & $95.4$ & $92.4$ & $86.7$ & $71.1$ \\
&
&
&
& Precision (\%) & $93.6$ & $92.5$ & $87.1$ & $72.1$ \\
\addlinespace

\multirow{2}{*}{Random}
& \multirow{2}{*}{$83.58$}
& \multirow{2}{*}{$1.12$}
& \multirow{2}{*}{$24.1 \pm 10.5$}
& Recall (\%)    & $18.8$ & $10.2$ & $4.8$ & $1.7$ \\
&
&
&
& Precision (\%) & $7.9$ & $7.3$ & $3.6$ & $1.2$ \\
\addlinespace

\multirow{2}{*}{BATS}
& \multirow{2}{*}{$87.23$}
& \multirow{2}{*}{$1.12$}
& \multirow{2}{*}{$23.9 \pm 11.2$}
& Recall (\%)    & $97.9$ & $95.8$ & $91.4$ & $77.1$ \\
&
&
&
& Precision (\%) & $73.8$ & $91.2$ & $85.2$ & $68.7$ \\

\bottomrule
\end{tabular}
\end{table*}

BATS achieves the highest segmentation accuracy in Table~5, reaching 87.23 HEC Dice compared with 84.55 for the iterative baseline and 83.58 for random placement. The iterative baseline matches or exceeds BATS in precision at every threshold, whereas its recall is lower at every split depth. Across these variants, the recall differences align more closely with the observed Dice scores than the precision differences: unnecessary fine tokens primarily increase computational cost, whereas a missed boundary region lacks the fine-resolution representation required by the refiner. The recall deficit widens from 2.5 points at $k=1$ to 6.0 points at $k=4$, consistent with errors propagating through sequential coarse-to-fine gating. The iterative baseline is also 26\% slower, requiring 30.2 rather than 23.9 seconds per volume, because its split levels are evaluated sequentially. Random placement reduces Dice by 3.65 points at the same token count, demonstrating that spatial placement contributes substantially beyond the token budget itself. The always-retained coarsest scale nevertheless preserves complete spatial coverage, limiting the severity of the degradation.

\subsubsection{Cross-scale interaction and component ablation.}

Table~\ref{tab:ablation_combined} reports two further ablations. Panel (a) compares cross-scale interaction mechanisms in the refiner: parent cluster attention improves over local cluster attention by $0.99$ Dice on KiTS and $0.42$ on AMOS, whereas supplying the same ancestor context through a separate cross-attention step recovers only part of this gain ($+0.08$ on KiTS). Because the variants have similar parameters and inference cost, this suggests that, under the evaluated settings, repeated ancestor injection inside every cluster-attention layer delivers context more effectively than a single dedicated pass, rather than adding capacity. 

Panel~(b) removes one component of the full model at a time. Removing raw-patch re-embedding leaves refiner tokens dependent only on boundary-predictor features; removing either auxiliary loss eliminates the corresponding composition-based supervision of the boundary predictor or intermediate refiner tokens; and removing oracle warmup uses the predicted token hierarchy from the start of training. Each removal reduces KiTS Dice by $0.73$--$1.02$ points, with oracle warmup producing the largest decrease.

\begin{table}[!ht]
\centering
\caption{Ablation studies. (a) Cross-scale interaction mechanism in the refiner, on KiTS and AMOS; all variants have similar parameters ($53$--$54$M) and inference cost. (b) Removing one component of the full model at a time, on a single KiTS fold. Values are mean (HEC) Dice.}
\label{tab:ablation_combined}
\footnotesize
\setlength{\tabcolsep}{3pt}
\vspace{3pt}
\begin{minipage}[t]{0.55\linewidth}
\centering
(a) Cross-scale interaction\\[2pt]
\begin{tabular}{lcc}
\toprule
Method & \makecell{KiTS\\HEC Dice} & \makecell{AMOS\\Dice} \\
\midrule
Cluster attention              & $86.24$ & $87.69$ \\
Separate parent cross-attn.\   & $86.32$ & $87.90$ \\
Parent cluster attn.\ (ours)   & $87.23$ & $88.11$ \\
\bottomrule
\end{tabular}
\end{minipage}\hfill
\begin{minipage}[t]{0.42\linewidth}
\centering
(b) Component removal (KiTS)\\[2pt]
\begin{tabular}{lc}
\toprule
Configuration & HEC Dice \\
\midrule
BATS (full)                & $87.23$ \\
No raw-patch re-embed.\    & $86.50$ \\
No boundary aux loss.\    & $86.28$ \\
No refiner aux loss.\      & $86.38$ \\
No oracle warmup           & $86.21$ \\
\bottomrule
\end{tabular}
\end{minipage}
\end{table}

\subsection{Qualitative analysis}

\begin{figure*}[!ht]
\centering
\begin{tabular}{@{}c@{\hspace{2pt}}
                    c@{\hspace{1pt}}
                    c@{\hspace{1pt}}
                    c@{\hspace{1pt}}
                    c@{\hspace{1pt}}
                    c@{}}
  & \small GT
  & \small Prediction
  & \makecell{\small Token\\[-1pt]\small selection}
  & \makecell{\small Zoomed\\[-1pt]\small prediction}
  & \makecell{\small Zoomed\\[-1pt]\small selection} \\[1pt]
  \rotatebox{90}{\small ACDC} &
  \qfull{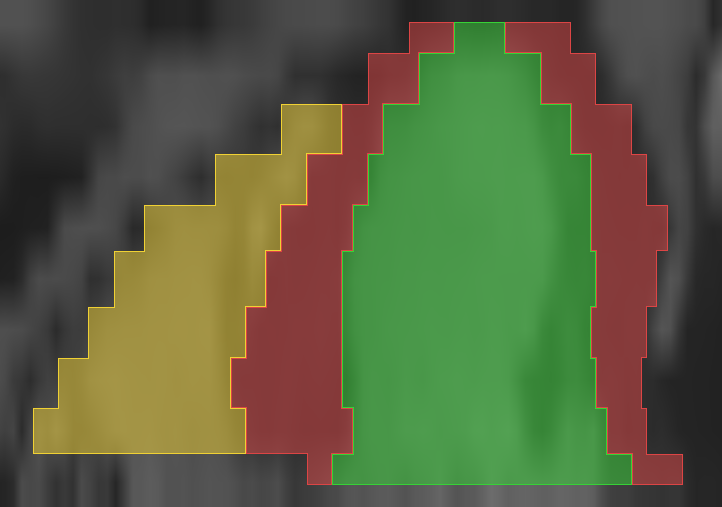} &
  \qfull{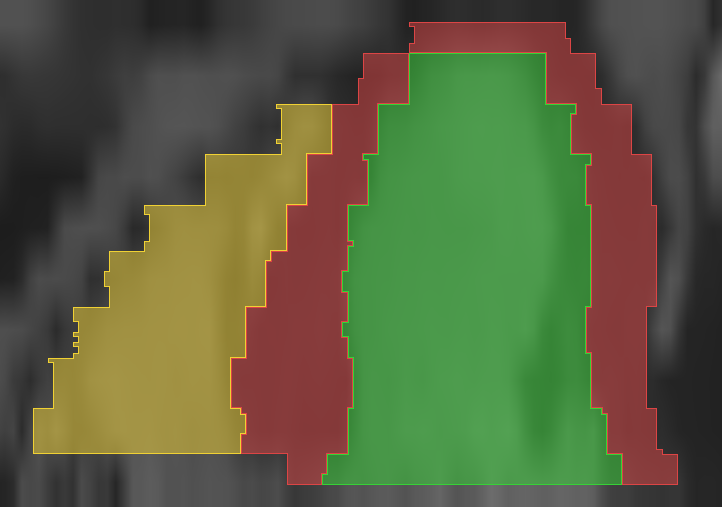} &
  \qfull{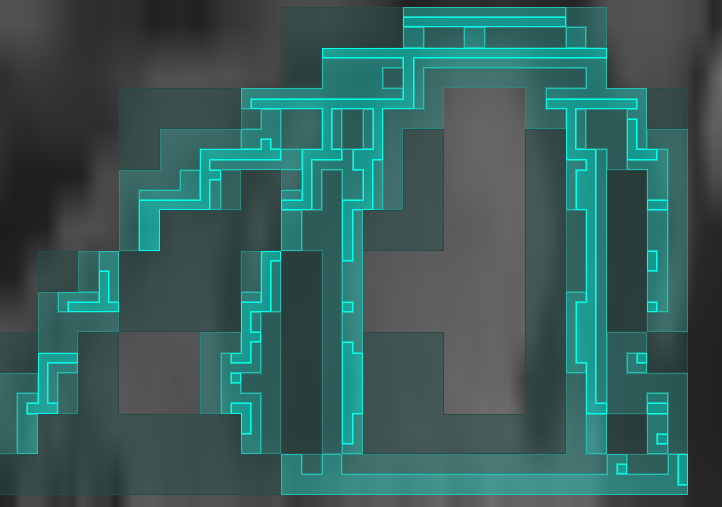} &
  \qzoom{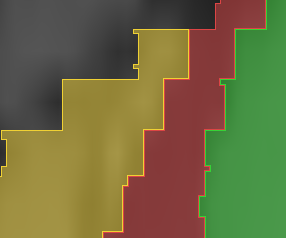} &
  \qzoom{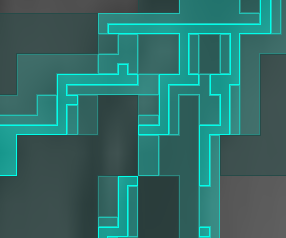} \\[2pt]
  \rotatebox{90}{\small LiTS} &
  \qfull{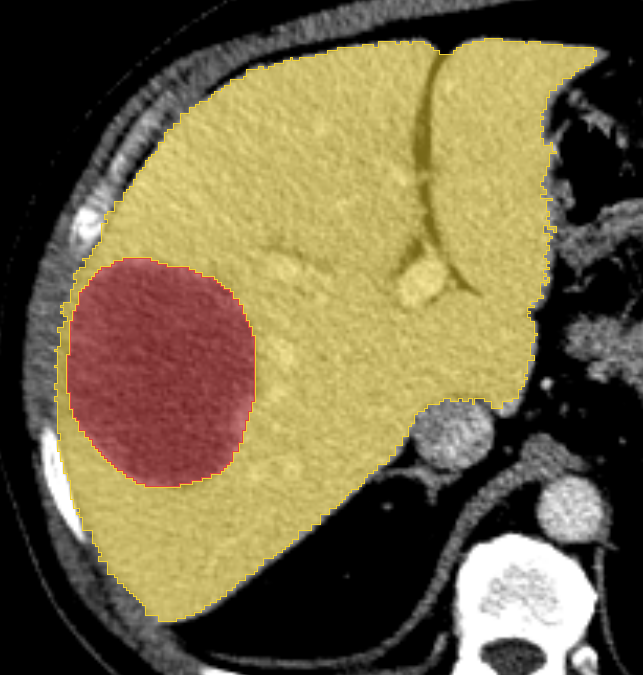} &
  \qfull{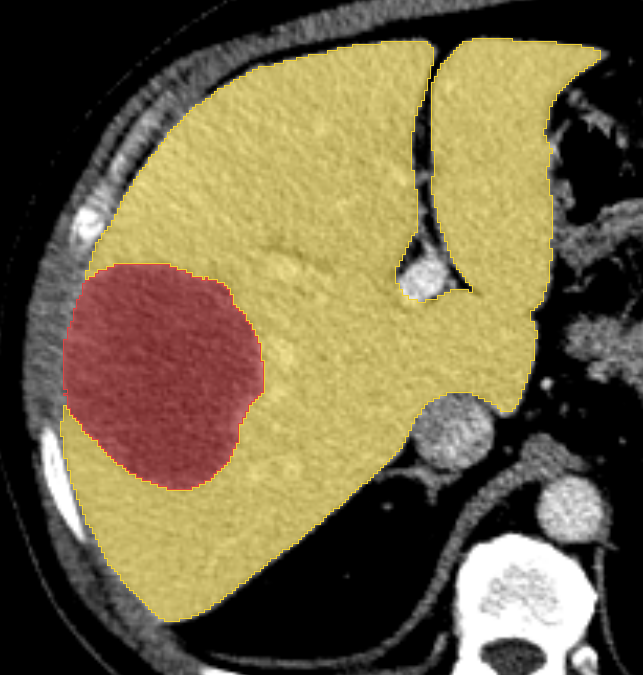} &
  \qfull{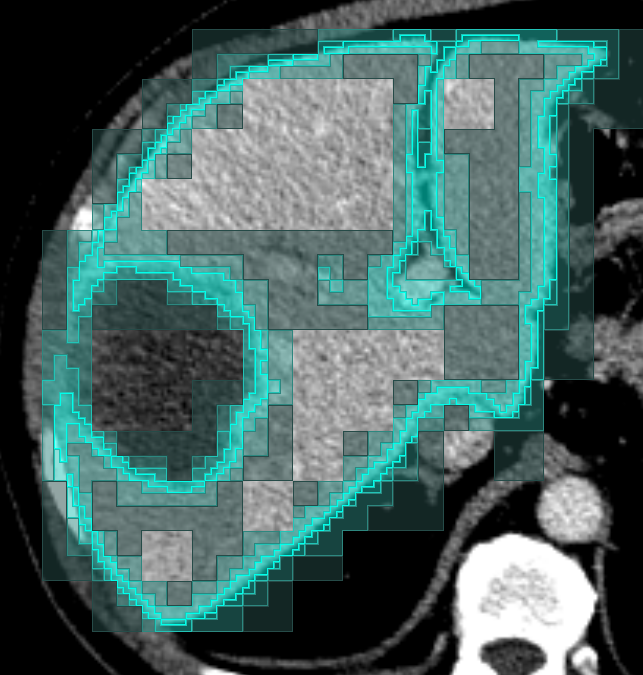} &
  \qzoom{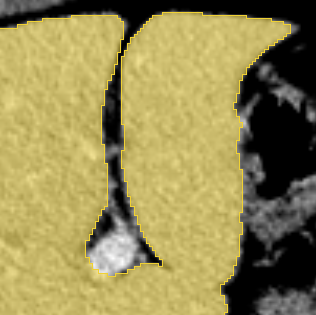} &
  \qzoom{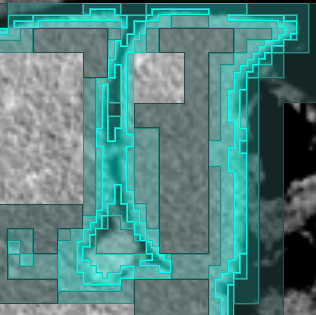} \\[2pt]
  \rotatebox{90}{\small BraTS} &
  \qfull{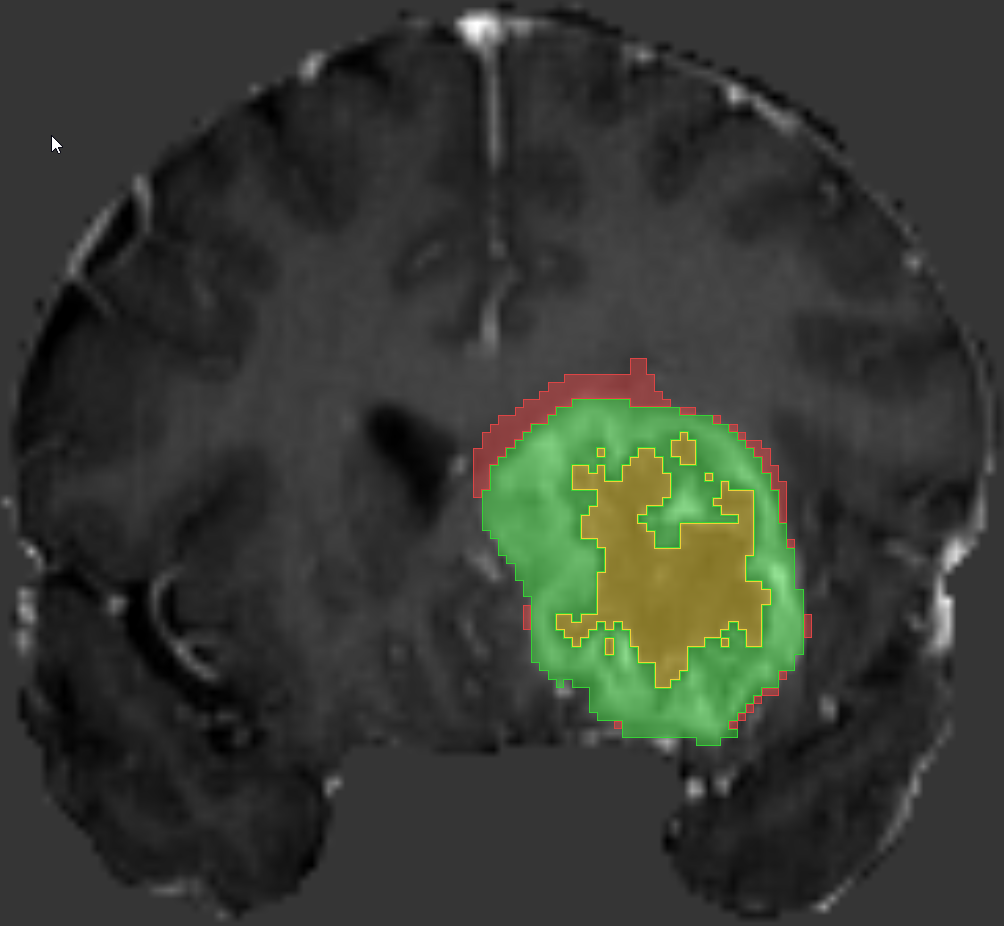} &
  \qfull{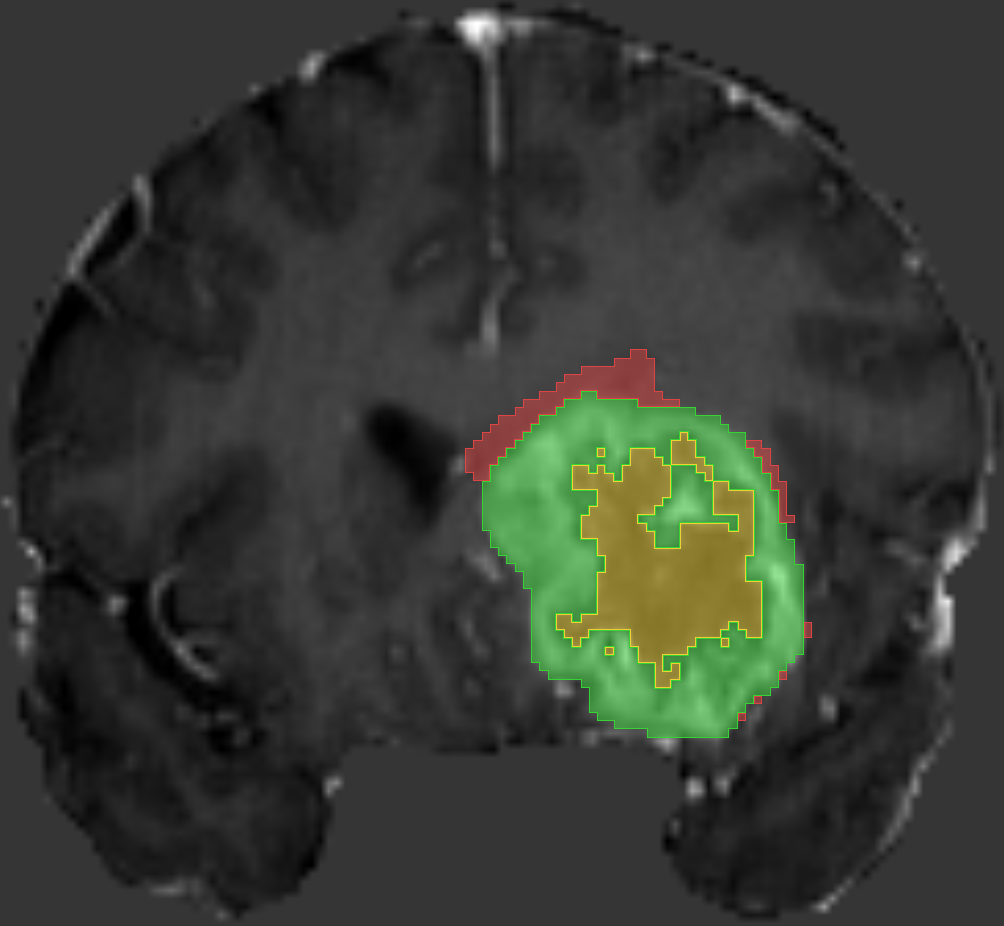} &
  \qfull{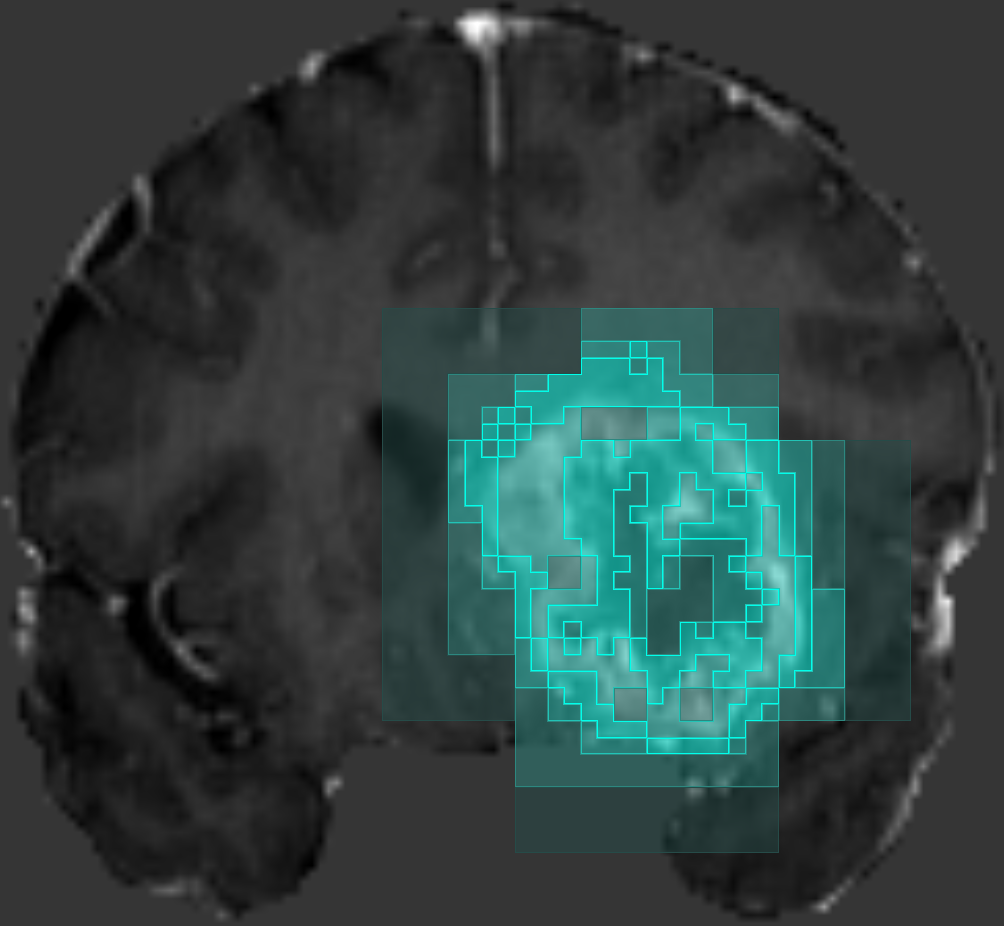} &
  \qzoom{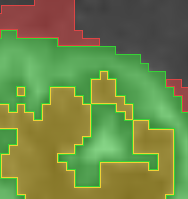} &
  \qzoom{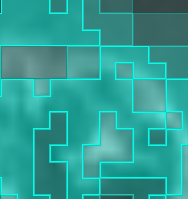} \\[2pt]
  \rotatebox{90}{\small KiTS} &
  \qfull{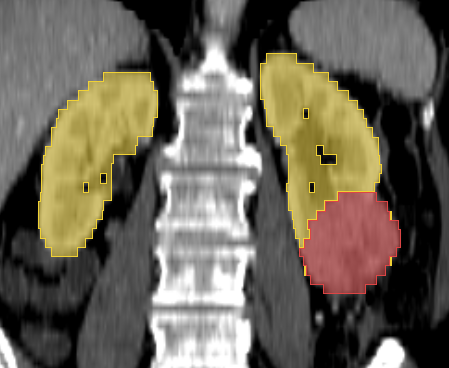} &
  \qfull{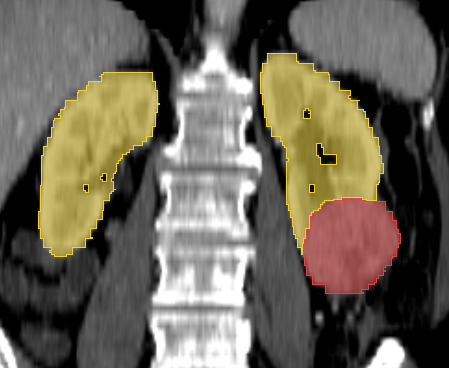} &
  \qfull{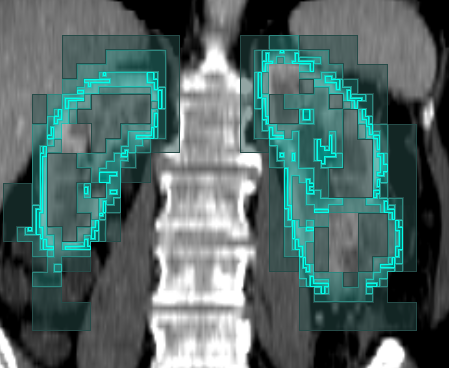} &
  \qzoom{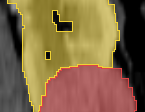} &
  \qzoom{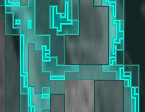} \\[2pt]
  \rotatebox{90}{\small AMOS} &
  \qfull{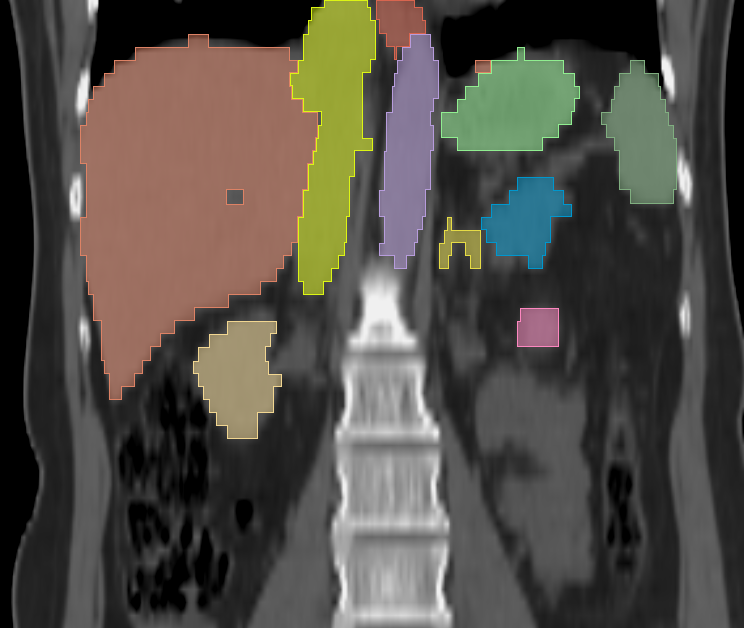} &
  \qfull{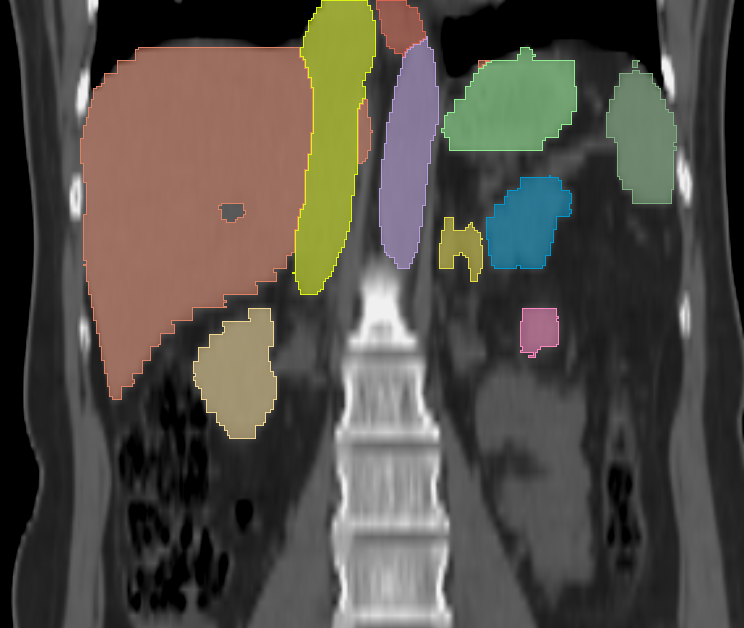} &
  \qfull{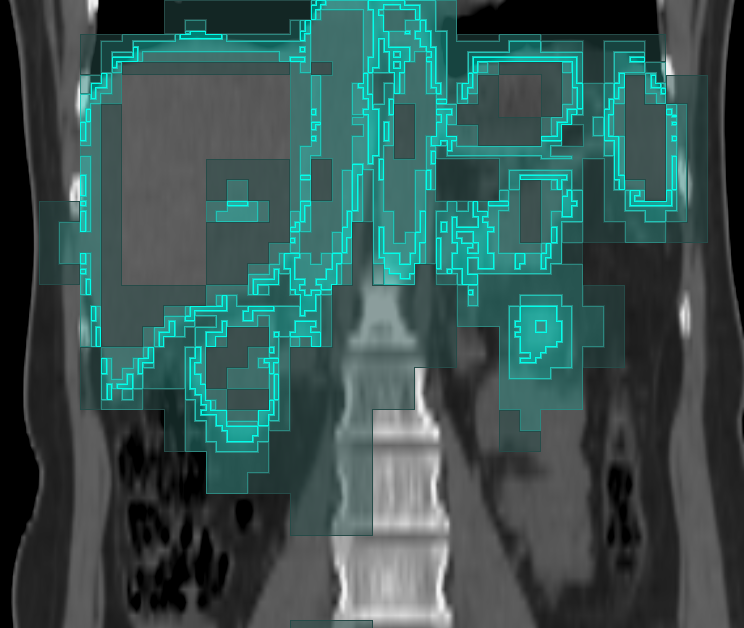} &
  \qzoom{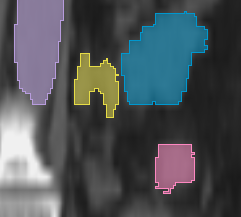} &
  \qzoom{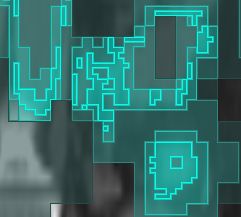}
\end{tabular}
\caption{Representative predictions and retained token hierarchies on ACDC, LiTS, BraTS, KiTS, and AMOS. From left to right: ground truth, prediction, token selection, zoomed prediction, and zoomed selection. Retained tokens are shown as filled teal patches, with darker colours indicating coarser tokens. Fine-resolution tokens generally follow class boundaries.}
\label{fig:qualitative_good}
\end{figure*}

Figure~\ref{fig:qualitative_good} shows one representative example per dataset. Predicted segmentations closely match the reference labels, and fine tokens concentrate near visible class boundaries while progressively coarser tokens cover homogeneous interiors and background. The examples span cardiac chambers, a tumour--kidney interface, brain-tumour subregions, and multiple abdominal organs, including the small left adrenal gland in the AMOS zoom. The LiTS example also contains fine tokens around a strong internal intensity discontinuity that is not a labelled semantic boundary, suggesting that boundary prediction responds to image contrast and can retain unnecessary fine tokens at non-semantic edges, consistent with the relatively low split-depth precision. Additional successful examples and stage-specific failure cases are provided in Appendix~\ref{app:qualitative}.

\section{Conclusion}

BATS uses input-dependent mixed-resolution tokens to reduce the inference cost of volumetric segmentation. Across five nnU-Net Revisited tasks it achieves the highest LiTS Dice and averages $0.37$ Dice points below MedNeXt-L k3, while reducing peak allocated memory by more than $53\%$ on KiTS, LiTS, and BraTS. Runtime improves by $27$--$30\%$ on the boundary-sparse KiTS and LiTS datasets but is higher than the dense baseline on BraTS, where BATS retains far more tokens. On the KiTS ablation fold, split-depth recall aligns more closely with segmentation accuracy than split-depth precision, and injecting ancestors into every attention layer outperforms a single cross-attention pass. These results support parallel boundary prediction with a recall-oriented fine-first context cascade, together with repeated ancestor injection during refinement. Adaptive resolution is therefore most effective when fine spatial structure occupies a limited fraction of the volume. We discuss limitations in Appendix~\ref{app:limitations}.

\section{Acknowledgments}
The computations were enabled by resources provided by the National Academic Infrastructure for Supercomputing in Sweden (NAISS), partially funded by the Swedish Research Council through grant agreement no. 2022-06725 as well as the Berzelius resource provided by the Knut and Alice Wallenberg Foundation at the National Supercomputer Centre.

\appendix

\section{Architecture and Training Details}
\label{app:impl}

\subsection{Architecture configuration.}
Table~\ref{tab:architecture_full} lists the complete per-scale configuration summarised in Table~\ref{tab:architecture}. Boundary-predictor blocks are 3D inverted-bottleneck ConvNeXt blocks with expansion ratios $(2,3,4,4)$; refiner blocks are parent cluster attention blocks with MLP ratio $4.0$, RoPE3D positional encoding, LayerScale ($10^{-4}$), and stochastic depth (drop-path $0.1$).

\begin{table}[!ht]
\centering
\caption{Full BATS architecture configuration, per scale. The full model has $53$M parameters.}
\label{tab:architecture_full}
\small
\setlength{\tabcolsep}{4pt}
\begin{tabular}{llrrrrrr}
\toprule
Component & \makecell{Params\\(M)} & \makecell[l]{Scale\\(patch)} & Width & Blocks & Heads & \makecell{Cluster\\size} & \makecell{Nbhd.\\size} \\
\midrule
\multirow{4}{*}{\makecell[l]{Boundary\\predictor}} & \multirow{4}{*}{$19$}
 & $16^3$ & 512  & 6 & -- & -- & -- \\
 & & $8^3$  & 256 & 8 & -- & -- & -- \\
 & & $4^3$  & 128 & 4 & -- & -- & -- \\
 & & $2^3$  & 64 & 3 & -- & -- & -- \\
\midrule
\multirow{5}{*}{Refiner} & \multirow{5}{*}{$34$}
 & $16^3$ & 512  & 8 & 16  & 8 & 48 \\
 & & $8^3$  & 256  & 8 & 8  & 8 & 40 \\
 & & $4^3$  & 128 & 8 & 4  & 8 & 32 \\
 & & $2^3$  & 64 & 4 & 2  & 8 & 24 \\
 & & $1^3$  & 32 & 2 & 1 & 8 & 16 \\
\midrule
\makecell[l]{Segmentation\\head} & $0.4$ & $1^3$ & 24 & 2 & -- & -- & -- \\
\bottomrule
\end{tabular}
\end{table}

\subsection{Boundary predictor.}
The boundary predictor is a bottom-up 3D ConvNeXt pyramid following the MedNeXt block design~\cite{roy2023mednext}. Each residual block contains a depthwise $3\times3\times3$ convolution, GroupNorm, and a $1\times1\times1$ expansion--compression pair with GELU; strided blocks perform downsampling. After the bottom-up pass, a top-down FPN fuses coarse context into each finer feature map: the coarser feature map is upsampled by nearest-neighbour interpolation, projected with a $1\times1\times1$ convolution, and added to the fine-scale features. Each scale then applies a boundary head, a pre-normalisation layer followed by two $1\times1\times1$ convolutions with GroupNorm and GELU in between, producing a per-location boundary logit. The split-level prediction maps are linked by a learned spatial residual gate: the coarser prediction map is upsampled trilinearly and added to the fine-scale prediction, modulated by $\sigma(\mathbf{g}_s)$ where $\mathbf{g}_s$ is a $1\times1\times1$ convolution on the fine-scale FPN features. Coarse evidence can thus reinforce fine predictions but is suppressed where fine evidence disagrees, and no map's evaluation is gated by a discrete split decision at a coarser scale.

\subsection{Refiner patch embedding.}
For each retained token at scale $s$ with patch size $p_s^3$ voxels, the raw intensities are gathered into a $p_s\times p_s\times p_s$ mini-volume and passed through a per-scale patch embedding. When $p_s > 1$, a $1\times1\times1$ stem convolution widens the input to a base width, followed by $\log_2(p_s)$ stride-2 ConvNeXt downsampling blocks with progressively doubled width that reduce the patch to a single vector, with a LayerNorm at the output. For $p_s = 1$ the conv path is skipped and each voxel is projected through a linear layer, GELU, and LayerNorm. The FPN feature $\mathbf{f}^{\text{FPN}}_s$ is gathered from the dense map at the token's grid position and projected to the model width with a per-scale linear layer and LayerNorm. The image gate logit $\gamma_s$ is a learnable scalar initialised so that $g_s\approx0.1$ at the start of training, so the fused feature initially relies on the FPN path.

\subsection{Parent cluster attention.}
Clusters are constructed independently per sample, and neighbourhoods index only tokens of the same sample, so attention never crosses sample boundaries. Because the number of retained tokens differs between samples, each batch is padded to a common length and padded slots are masked out of every softmax; ancestor lookups index the hash grid by $(\text{batch}, \text{scale}, x, y, z)$, keeping resolution within one sample. In a block processing all $N_s$ scales, a finest-scale token receives $N_s-1$ ancestor injections and coarsest-scale tokens receive none, running standard cluster attention. Although each token accesses only its direct ancestors, each ancestor has itself attended over a neighbourhood at its coarser scale, so the effective context window exceeds the local cluster. The space-filling ordering used to form balanced clusters follows AutoFocusFormer~\cite{ziwen_autofocusformer_2023}.

\subsection{Auxiliary supervision.}
The boundary-predictor auxiliary head reads the boundary predictor's features at each split level and predicts the class-fraction distribution of the corresponding patch. It is trained with soft cross-entropy against the ground-truth per-patch class fractions and evaluated only at positions retained by the fine-first context cascade. It does not participate directly in the boundary-prediction decisions: it does not change which patches split and only shapes the shared features. The refiner auxiliary head is a per-token head at each refiner scale that predicts the class composition of the token's patch, trained with a soft cross-entropy summed over scales, so coarse tokens receive a direct segmentation-shaped signal.

\subsection{Optimisation and preprocessing.}
The total loss is a weighted sum of the segmentation loss ($1.0$), boundary-prediction loss ($1.0$), and the two auxiliary losses ($0.15$ each). All images are resampled to a $1\times1\times1$~mm voxel spacing. BATS uses the patch hierarchy $\{16,8,4,2,1\}$, boundary threshold $\tau=0.5$, and oracle warmup fraction $25\%$. The number of training epochs, augmentations, and preprocessing are taken unchanged from MedNeXt~\cite{roy2023mednext,isensee2024nnu}.

\section{Evaluation Details}

\subsection{Efficiency measurement protocol.}
\label{app:efficiency}

Window stitching is performed on the CPU. We exclude the first case from the memory and time measurements, since MedNeXt runs a cuDNN benchmarking pass on the first case that transiently inflates peak memory and runtime; this case also serves as a warm-up. Peak memory is read from torch with \texttt{max\_memory\_allocated} after reseting memory stats, and per-volume timing brackets the full sliding-window inference with torch \texttt{synchronize} before and after each case. The large spread in per-volume inference time reflects the widely varying volume sizes, since the number of sliding windows grows with volume extent.

\subsection{Boundary-prediction metrics and analysis.}
\label{app:selection_metrics}
The $N_s=5$ token scales consist of one coarsest scale that is always retained and $N_s-1=4$ finer scales introduced by split decisions. For each voxel $v$, the ground-truth split depth $k^*_v\in\{0,\ldots,N_s-1\}$ counts how many split decisions along the path from the coarsest token to $v$ are positive; a decision is positive if the corresponding parent patch contains more than one semantic class. The predicted split depth $\hat{k}_v$ counts how many split decisions along that path result in a retained finer token covering $v$. At each threshold $k\in\{1,\ldots,N_s-1\}$, a voxel is a positive if its ground-truth split depth is at least $k$, and we report precision and recall. Low recall at threshold $k$ indicates boundary regions not selected at sufficient resolution; low precision indicates non-boundary regions selected at unnecessarily fine resolution.

Beyond token count and placement, an asymmetry in supervision likely contributes to the widening recall gap of the iterative boundary-prediction baseline. In the iterative design, the finer-scale boundary heads receive gradient only through positions that passed the earlier thresholds, whereas BATS supervises every split-level head over the full crop. The same asymmetry propagates through the shared feature pipeline: a position missed by the iterative boundary predictor receives no refiner gradient either, so its fine-scale features cannot improve, making that region persistently harder to recover in later epochs. Dense supervision at all split levels avoids this feedback loop. BATS's low coarse-scale precision ($P_{\geq1}=0.738$ vs.\ $0.936$) is a direct consequence of the fine-first context cascade: retaining every selected token's ancestors trades precision for recall at all but the finest level, most strongly at the coarsest scale because retaining any descendant also requires retaining its ancestors.

\section{Qualitative Results}
\label{app:qualitative}

\begin{figure*}[!ht]
\centering
\begin{tabular}{@{}c@{\hspace{2pt}}
                    c@{\hspace{1pt}}
                    c@{\hspace{1pt}}
                    c@{\hspace{1pt}}
                    c@{\hspace{1pt}}
                    c@{}}
  & \small GT
  & \small Prediction
  & \makecell{\small GT-based\\[-1pt]\small selection}
  & \makecell{\small Predicted\\[-1pt]\small selection}
  & \makecell{\small Failure\\[-1pt]\small zoom} \\[1pt]

  \rotatebox{90}{\small LiTS} &
  \qfull{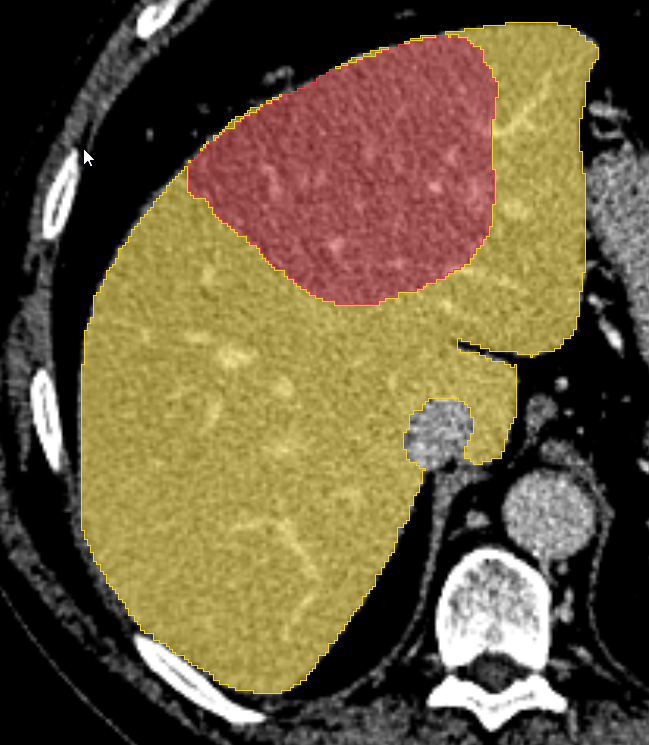} &
  \qfull{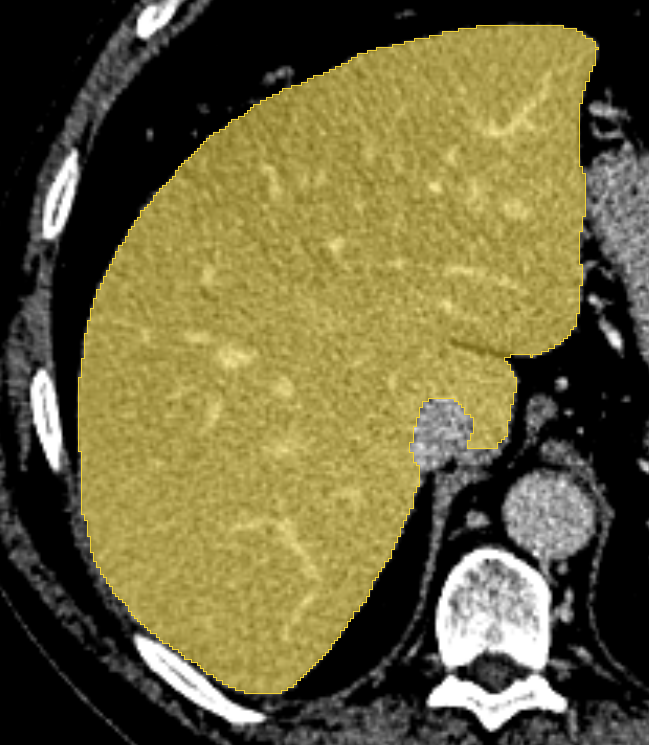} &
  \qfull{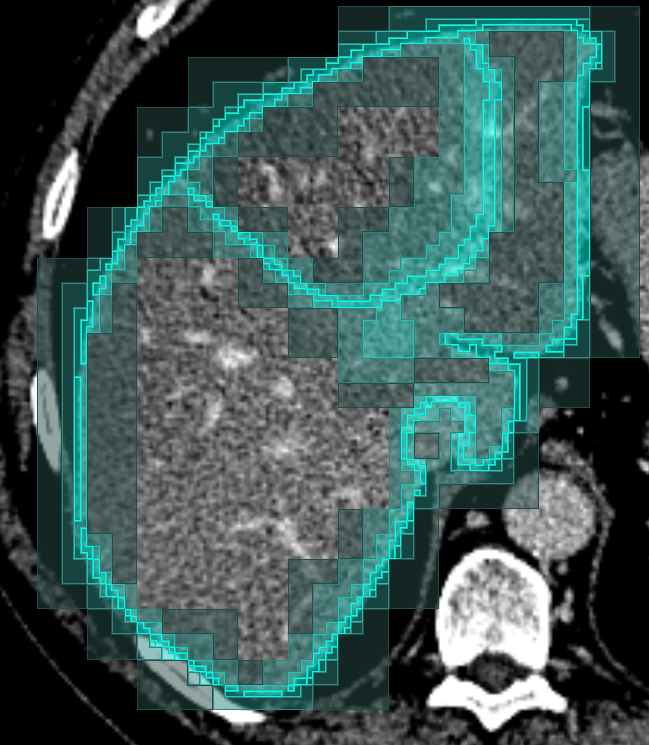} &
  \qfull{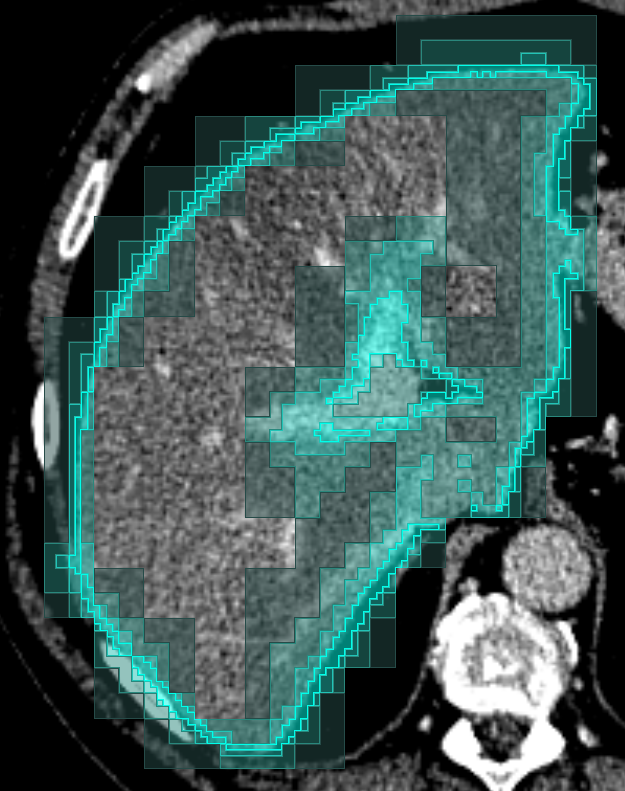} &
  \qzoom{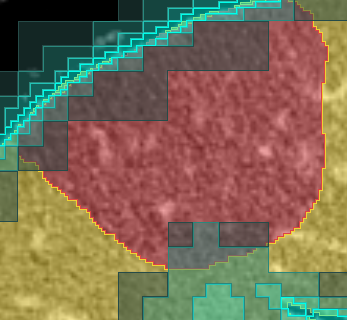} \\

  \rotatebox{90}{\small KiTS} &
  \qfull{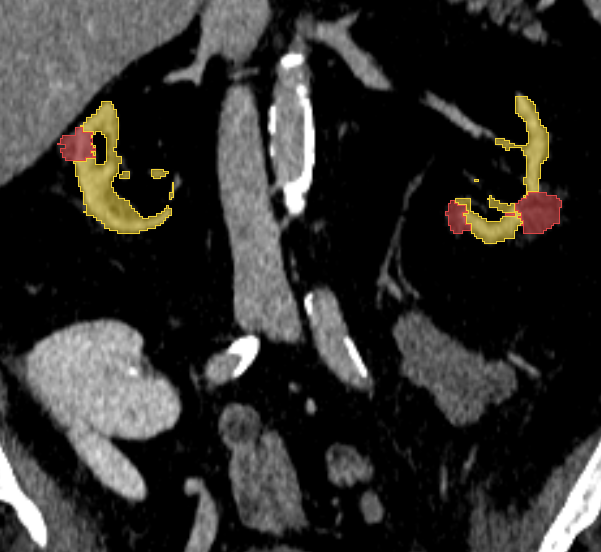} &
  \qfull{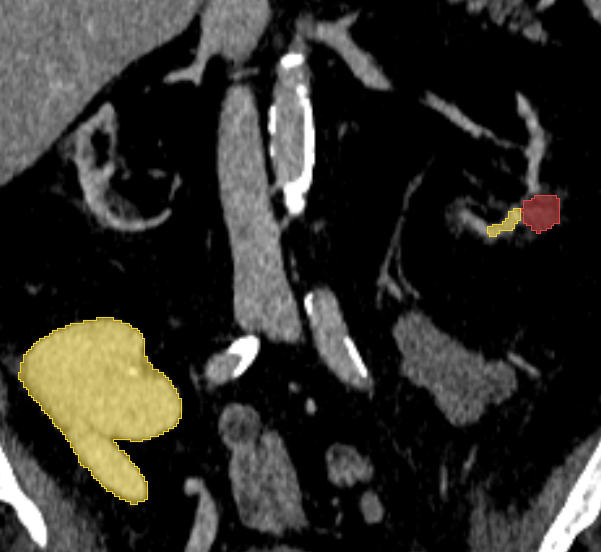} &
  \qfull{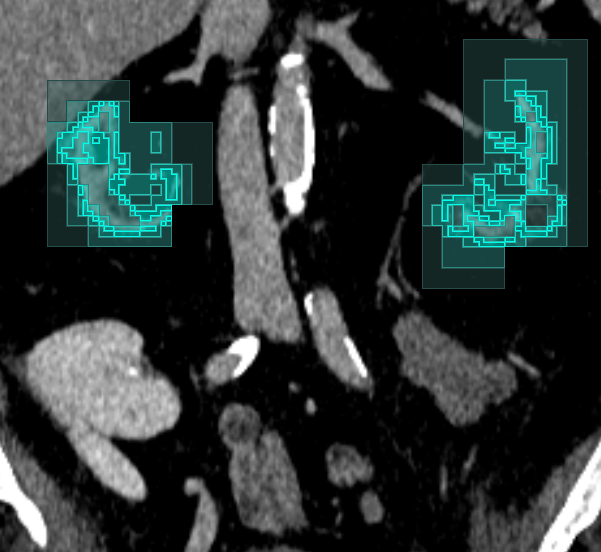} &
  \qfull{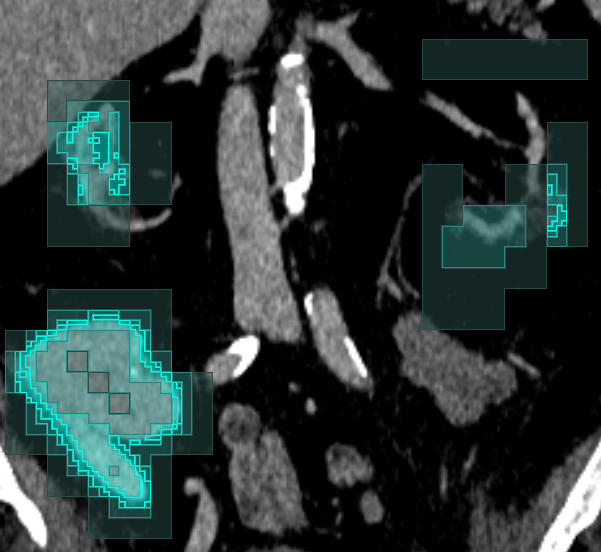} &
  \qzoom{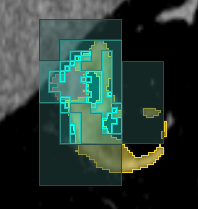} \\

  \rotatebox{90}{\small AMOS} &
  \qfull{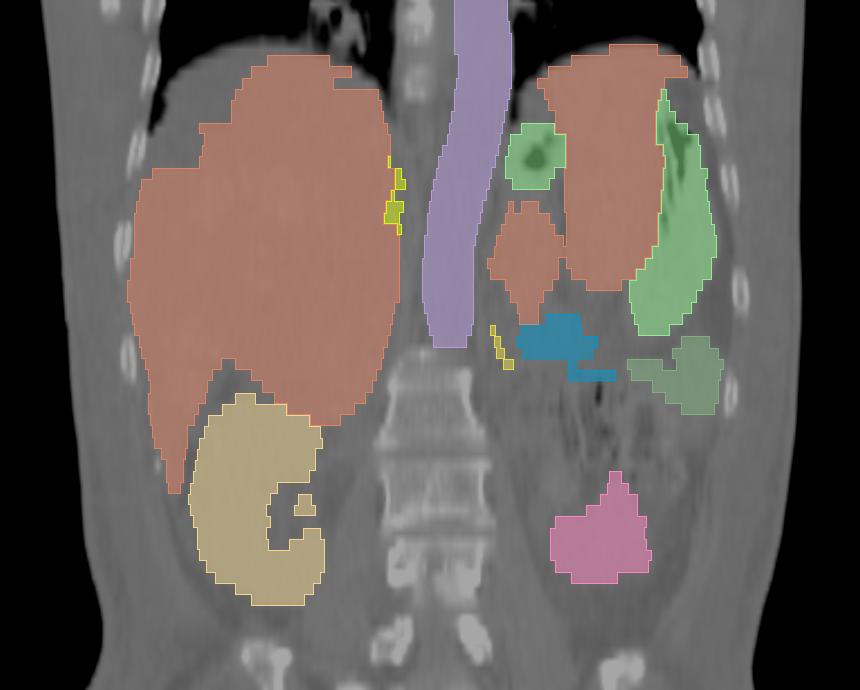} &
  \qfull{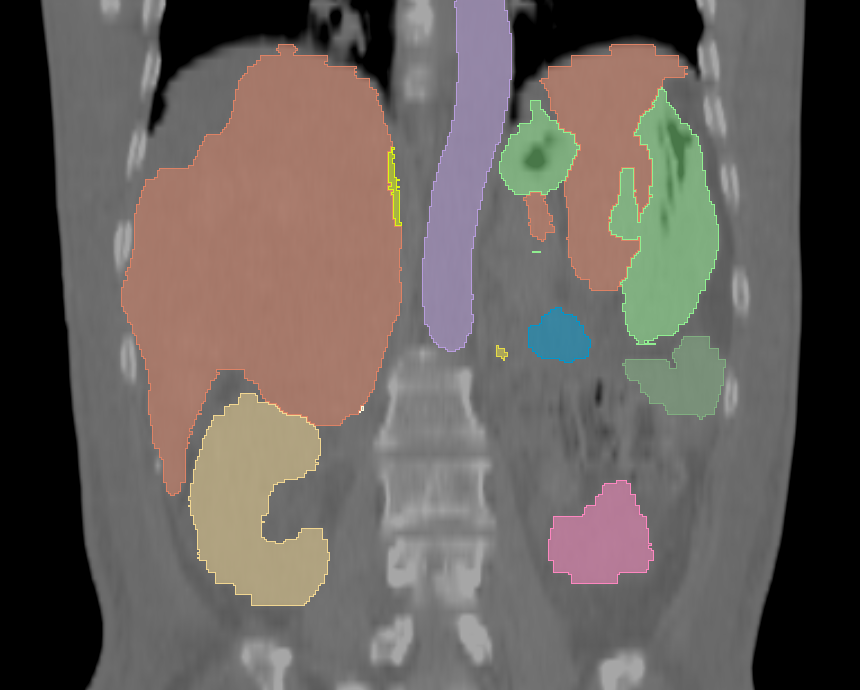} &
  \qfull{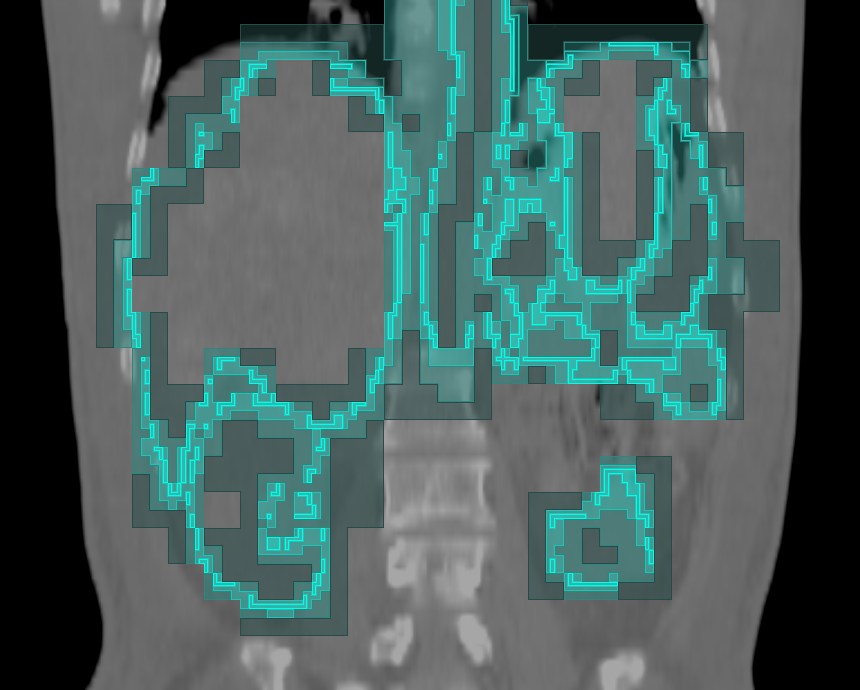} &
  \qfull{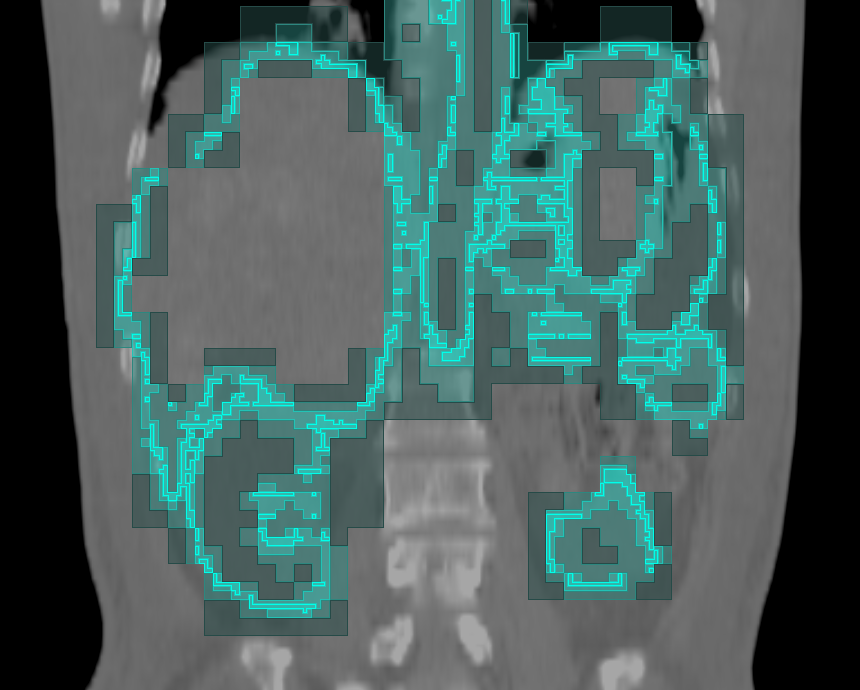} &
  \qzoom{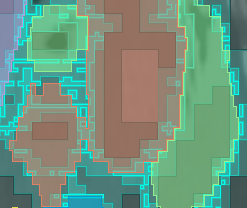}
\end{tabular}

\caption{Failure cases on LiTS, KiTS, and AMOS. From left to right:
ground truth, prediction, GT-based token selection, predicted token
selection, and a zoomed crop showing the ground-truth segmentation over
the predicted selection. Token colours follow
Figure~\ref{fig:qualitative_good}.}
\label{fig:qualitative_failure}
\end{figure*}

Figure~\ref{fig:qualitative_failure} illustrates failures arising at different stages. In the LiTS example, boundary prediction fails to select sufficient fine-resolution tokens around the missed tumour, leaving the refiner without the corresponding fine-scale representation. The KiTS example combines boundary-prediction and classification errors: fine tokens cover parts of the missed structures, while other relevant boundaries receive insufficient resolution, and several selected regions are still misclassified. In the AMOS example, the predicted token selection closely follows the reference boundary pattern in the failed region, but the liver is partly classified as stomach or background, indicating that the error occurs after construction of the token hierarchy. These cases show that missed boundary predictions can remove fine-scale information irreversibly, whereas an appropriate token selection does not by itself guarantee correct semantic classification.

\section{Limitations.}
\label{app:limitations}
The efficiency evaluation is limited to three datasets, one principal dense comparator, and a single hardware and software configuration. Wall-clock performance of sparse operators may vary across devices and implementations, and boundary-aware mixed-resolution processing may provide a smaller advantage for targets whose discrimination depends on diffuse texture rather than localised spatial structure. Future work should evaluate the architecture across additional hardware and clinical tasks, improve calibration of the boundary-prediction accuracy--cost trade-off, and investigate whether further sparsification of the dense predictor and segmentation head can yield additional reductions in memory and latency.

\nocite{ulen-tmi-2013}

\bibliographystyle{splncs04}
\bibliography{references}

\end{document}